\documentclass[preprint,12pt]{elsarticle}

\usepackage{amssymb}

\usepackage{amsmath}
\usepackage{hyperref}
\usepackage{booktabs}
\usepackage{bm} 
\usepackage[caption=false,font=normalsize]{subfig}
\usepackage{multicol} 
\usepackage{multirow}
\usepackage{svg}
\usepackage{soul}

\newcommand{\ve}[1]{\mathbf{#1}}
\usepackage[table]{xcolor} 
\definecolor{lightpurple}{RGB}{229,204,255} 

\journal{Neurocomputing}

\begin{document}

\begin{frontmatter}



\title{Class Incremental Learning with Task-Specific Batch Normalization and Out-of-Distribution Detection}



\author[label1,label3]{Zhiping Zhou}
\author[label1,label3]{Xuchen Xie}
\author[label4]{Yiqiao Qiu\fnref{equal,unrelated}}
\author[label1,label3]{Run Lin\fnref{equal}} 
\author[label1,label3]{Wei-Shi Zheng}
\author[label1,label2,label3]{Ruixuan Wang\corref{corresp}}

\cortext[corresp]{Corresponding author: wangruix5@mail.sysu.edu.cn}
\fntext[equal]{Authors contributed equally.}
\fntext[unrelated]{Works not related to role at Amazon.}

\affiliation[label1]{organization={School of Computer Science and Engineering, Sun Yat-sen University},
            city={Guangzhou},
            state={Guangdong},
            country={China}}

\affiliation[label2]{organization={Peng Cheng Laboratory},
            city={Shenzhen},
            state={Guangdong},
            country={China}}

\affiliation[label3]{organization={Key Laboratory of Machine Intelligence and Advanced Computing, MOE},
            city={Guangzhou},
            state={Guangdong},
            country={China}}

\affiliation[label4]{organization={Department of Computer Science and Engineering, University of California San Diego},
            city={San Diego},
            state={California},
            country={United States}}

\begin{abstract}
This study focuses on incremental learning for image classification, exploring how to reduce catastrophic forgetting of all learned knowledge when access to old data is restricted. The challenge lies in balancing plasticity (learning new knowledge) and stability (retaining old knowledge). Based on whether the task identifier (task-ID) is available during testing, incremental learning is divided into task incremental learning (TIL) and class incremental learning (CIL). The TIL paradigm often uses multiple classifier heads, selecting the corresponding head based on the task-ID. Since the CIL paradigm cannot access task-ID, methods originally developed for TIL require explicit task-ID prediction to bridge this gap and enable their adaptation to the CIL paradigm. {In this study, a novel continual learning framework extends the TIL method for CIL by introducing out-of-distribution detection for task-ID prediction. Our framework utilizes task-specific Batch Normalization (BN) and task-specific classification heads to effectively adjust feature map distributions for each task, enhancing plasticity. With far fewer parameters than convolutional kernels, task-specific BN helps minimize parameter growth, preserving stability. Based on multiple task-specific classification heads, we introduce an ``unknow'' class for each head. During training, data from other tasks are mapped to this unknown class. During inference, the task-ID is predicted by selecting the classification head with the lowest probability assigned to the unknown class.} Our method achieves state-of-the-art performance on two medical image datasets and two natural image datasets. {The source code is available at \url{https://github.com/z1968357787/mbn_ood_git_main}.}
\end{abstract}



\begin{keyword}


Task Incremental Learning \sep  Class Incremental Learning\sep Batch Normalization\sep Out-of-Distribution Detection
\end{keyword}

\end{frontmatter}



\section{Introduction}
\label{sec:intro}

{Deep learning models have been extensively applied across various domains, including image recognition~\cite{nc1,nc2}, natural language processing~\cite{nc3,nc4}, generative adversarial networks~\cite{intro1,intro4}, and industrial applications such as multimodal arc detection~\cite{intro2,intro3}.}
However, they usually rely on large amounts of data for training to achieve high performance. 
In real-world scenarios, it is often difficult to obtain data for all necessary classes at once, leading to data arriving in multiple phases. Moreover, the classes appearing in different phases are typically non-overlapping. Consequently, AI models often need to learn new classes gradually. 
With privacy and memory constraints, models can only access data from the current phase and are restricted to retrieve data from previous phases. 
Research has found that under these circumstances, models are susceptible to catastrophic forgetting~\cite{french1999catastrophic}, a phenomenon characterized by a substantial decline in performance on previously learned tasks.
Therefore, enabling models to continually learn new knowledge without forgetting old knowledge is one key challenge for promoting the widespread application of deep learning based AI across various fields.

Multiple approaches have been proposed to alleviate catastrophic forgetting. However, they often struggle to balance the model's stability (ability to retain old knowledge) and plasticity (ability to learn new knowledge), and may also face challenges in controlling parameter growth. For instance, methods based on model parameter regularization~\cite{ewc,crnet,InterContiNet} aim to protect crucial neural network weights from being updated drastically through constraints, thus preventing old knowledge loss. However, this protection can lead to model rigidity, making it difficult for the model to assimilate new knowledge. Another type of method is the dynamic expansion of model structure~\cite{pnn,dytox,DER,acl}, where new layers or sub-modules are added for new tasks. Although such methods significantly enhance model plasticity, they inevitably increase memory demands. 
Therefore, effectively adding network components without substantially increasing the memory burden is a fundamental issue for this type of method. Overall, existing methods struggle to achieve a good balance between plasticity, stability and memory control, which is the core challenge in incremental learning.

In a regular sequential multi-task learning process (without incremental learning), a single feature extractor without any restriction is used to fit the distinctive knowledge representations of each task sequentially. During this process, the feature extractor focuses on learning new tasks only, leading to catastrophic forgetting of old tasks. Minimizing updates of the feature extractor can substantially mitigate or completely avoid catastrophic forgetting. {In this paper, we propose a method that minimizes changes to learned feature representations by introducing task-specific Batch Normalization (BN) layers and classification heads. Since BN layers contain far fewer parameters than convolutional layers in typical CNNs, they can be added for each task with minimal parameter growth. These task-specific modules are trained and then preserved in subsequent continual learning, which helps mitigate catastrophic forgetting and maintain model stability. Moreover, BN effectively reshapes feature distributions, enhancing the model’s ability to capture task-specific characteristics and improving learning plasticity. Because the task identifier (task-ID) is unavailable in the class-incremental learning (CIL) setting, an additional mechanism is required at inference time to select the appropriate task-specific modules. For each task-specific BN layer and classification head, data from other tasks can be regarded as out-of-distribution (OOD) samples. To leverage this property, we augment each classification head with an ``unknown'' class. During training, samples from other tasks are mapped to this class, enabling the classification heads to perform implicit OOD detection. During inference, the task-ID is predicted by selecting the head with the lowest probability assigned to the unknown class, allowing the model to choose the most appropriate task-specific modules.}

This work extends our prior conference publication~\cite{mbn_til} where we made two foundational contributions:
\begin{enumerate}
    \item Pioneered task-specific BN in Task-Incremental Learning (TIL);
    \item Established theoretical guarantees. Under task-ID availability, we have demonstrated that task-specific BN layers and classifier heads can significantly enhance within-task classification performance.
\end{enumerate}

Building upon these foundations, this study includes the following major contributions:

\begin{enumerate}
    \item We introduce task-specific BN in the CIL paradigm for the first time.

    \item Based on multiple task-specific classifier heads, we introduce out-of-distribution detection for task-ID prediction, demonstrating that a task-ID prediction mechanism can effectively extend TIL to the CIL setting, where the task-ID is unknown.

    \item Our proposed method achieves state-of-the-art CIL performance on two medical image datasets and two natural image datasets, 
    achieving a better balance among model stability,  plasticity, and parameter growth.
\end{enumerate}

\section{Related Work}

\label{sec:formatting}
\subsection{General Overview}
Incremental learning for image classification is divided into two main paradigms, namely, task incremental learning~\cite{til2018_1, til2019_2} (TIL) and class incremental learning~\cite{til2018_1,til2019_2} (CIL). {Both TIL and CIL learn new class knowledge sequentially in a task-based manner during training. Each task contains multiple classes, and the class sets across tasks are mutually disjoint. During testing, TIL provides the task identity for each sample, commonly referred to as the task identifier (task-ID), whereas CIL does not provide this information.} 
The CIL methods can be transferred to TIL by focusing only on specific categories associated with the task-ID during testing. However, adapting methods from TIL to CIL remains a significant challenge, mainly because choosing the correct classifier head from multiple options to make the final prediction becomes particularly challenging without task-ID. Thus, for most existing methods, the main difference between these two paradigms lies in that the TIL paradigm has multiple classifier heads specific to certain tasks, while the CIL paradigm typically has only one classifier head that adapts to all tasks. CIL does not require task-ID as prior knowledge, making it more aligned with practical applications. Therefore, this paper focuses on CIL and introduces methods related to CIL.

\subsection{Class Incremental Learning (CIL)}  

Many methods have been proposed to directly address the catastrophic forgetting problem in the CIL domain~\cite{DER,yangyang,nc_cl1,nc_cl2,nc_cl3,nc_cl4}.
{Most existing methods can be categorized into five types~\cite{cil_survey,cil_pre_trained_survey}.}


{The first type is regularization-based, such as CRNet~\cite{crnet} and InterContiNet~\cite{InterContiNet}, which introduce importance-aware regularization terms into the loss function to restrict parameter updates during the learning of new tasks, thereby preventing significant changes to parameters that are crucial for previous tasks. However, such methods inherently constrain model updates, which limits the model’s learning capacity and often hinders performance on new tasks.}

{The second type is distillation-based methods, which transfer knowledge from previous models to newly learned ones by encouraging similar feature representations. Representative approaches include LwF~\cite{lwf}, PODNet~\cite{podnet}, WA~\cite{wa}, UCIR~\cite{ucir}, and DSGD~\cite{dsgd}. These methods typically impose a distillation loss between corresponding layers (e.g., feature maps or logits) to preserve prior knowledge and mitigate catastrophic forgetting. Some studies further explore task-agnostic distillation~\cite{review4_1,review4_2,review4_3}, whose objective is not immediate task-specific performance but the development of a stable and general knowledge base through broad, self-driven knowledge acquisition across tasks without explicit task supervision.}

The third type of methods is replay-based, which alleviates model forgetting by replaying a small portion of real or synthetic old samples, including iCaRL~\cite{icarl}, DER++~\cite{darker}, DRC~\cite{drc}, CREATE~\cite{create}, and VBM-TCIL~\cite{vbm}. Most existing methods also rely on replay, such as knowledge distillation-based methods, which typically require replaying old samples to enhance the distillation effects. However, in incremental learning paradigms where tasks are presented with infinite data streams, storage and memory are always limited. Therefore, how to effectively select and utilize the representative samples for old sample replay is worth exploring.

{The fourth type is expansion-based methods, which progressively enlarge the model to increase representational capacity, such as PNN~\cite{pnn} and DyTox~\cite{dytox}. A representative example is DynaER~\cite{DER}, which introduces new learnable feature extractors as new tasks arrive to incorporate additional feature dimensions. These methods are often combined with regularization-based approaches~\cite{review4_1,review4_4}, which explicitly constrain updates to parameters identified as important within previously learned modules to preserve consolidated knowledge. They are also frequently integrated with replay-based strategies~\cite{review4_5} that leverage the experience encoded in old task modules to sample representative training data from previous tasks, enabling joint training of new task modules and mitigating catastrophic forgetting.}

The fifth type of methods is pretrained model-based, which leverage powerful pretrained backbones such as ViT~\cite{vit}, ResNet~\cite{resnet}, and CLIP~\cite{CLIP} to enhance continual learning performance. Representative methods include L2P~\cite{l2p}, MG-CLIP~\cite{mg_clip}, Engine~\cite{engine}, AttriCLIP~\cite{wang2023attriclip}, and PROOF~\cite{zhou2025learning}, among others. These methods typically keep the pretrained backbone frozen while inserting a small number of learnable parameters for fine-tuning, such as prompts~\cite{l2p,wang2023attriclip} or adapters~\cite{engine,zhou2025learning}. This design allows the model to fully exploit the knowledge embedded in the pretrained backbone while mitigating forgetting during new task learning, improving the balance between knowledge retention and acquisition.

{Our method belongs to pretrained model-based continual learning approaches and can also be viewed as a task-ID prediction-based framework. Similar to existing pretrained model-based methods, we freeze the pretrained backbone to preserve the rich knowledge acquired during pretraining. To enable efficient adaptation to new tasks, we introduce task-specific modules for each task, allowing the model to rapidly capture task-relevant knowledge while mitigating interference with previously learned representations.} 
Compared with methods that accumulate prompts~\cite{l2p,wang2023attriclip} or adapters~\cite{zhou2025learning,engine}, our approach only requires training independent BN layers and the classification head for each new task. This design preserves the strong feature representations of the pretrained backbone while substantially reducing parameter growth and computational overhead. Furthermore, by introducing an ``unknow'' class and performing OOD detection alignment, our method can effectively identify task-ID. With only a small amount of memory replay, it achieves strong resistance to forgetting, offering clear practical advantages in task-agnostic continual learning settings.

\subsection{Batch Normalization in Continual Learning} 
{Batch Normalization (BN) plays a crucial role in deep neural networks by normalizing layer inputs, which significantly reduces internal covariate shift, accelerates convergence, and improves training stability. Due to its ability to stabilize feature distributions and accelerate convergence, BN has been recently explored in continual learning to mitigate catastrophic forgetting. 

Recent studies~\cite{NBN,AdaB2N,TBBN,CN} have proposed various adaptations of BN to alleviate this issue. For example, NBN~\cite{NBN} injects Gaussian noise into BN outputs during training to simulate the statistical variations of future tasks, forcing higher-level layers to adapt to feature distribution shifts and improving robustness against distributional changes. AdaB2N~\cite{AdaB2N} introduces Bayesian learnable weights during training to dynamically balance the contributions of different tasks to normalization statistics, thereby enhancing the gradient–representation correlation and training stability. At inference, it employs an adaptive momentum adjustment mechanism to balance historical and current statistics, thereby reducing bias toward recent tasks. TBBN~\cite{TBBN} constructs task-balanced virtual batches by dynamically reshaping tensors, computes unbiased statistics, and modifies gradient backpropagation so that the BN parameters are updated under a balanced task state. At inference, it leverages statistics closer to the global distribution, thereby stabilizing features and reducing forgetting. CN~\cite{CN} concatenates Group Normalization (GN) with BN, where GN first performs spatial normalization to reduce dependence on global statistics, followed by affine-free BN to encourage knowledge sharing. This reduces reliance on running statistics during inference and achieves a balance between mitigating forgetting and promoting learning. }


Our method shares a similar motivation with existing approaches that modify BN layers, aiming to enable the model to distinguish distributional differences across tasks and thereby alleviate catastrophic forgetting in continual learning. However, unlike prior methods, we explicitly adopt task-specific BN layers to independently learn and preserve distributional knowledge for each task.
TSN-IQA~\cite{TSN-IQA} also extends BN to task-specific BN in continual blind image quality assessment (BIQA), while freezing the convolutional backbone, thus preserving old-task distributions. During inference, a gating module dynamically fuses the outputs of different BN branches, effectively reducing forgetting caused by normalization bias.
In contrast, while TSN-IQA focuses on learning across streams of image distortion types (e.g., JPEG compression and Gaussian blur), our method applies task-specific BN within a CIL framework. Specifically, the task-specific BN parameters for each task are optimized to capture and adapt to the feature distributions of their corresponding object categories, thereby mitigating catastrophic forgetting in CIL. Further distinctions between our method and TSN-IQA are discussed in the Method section.
\color{black}

\subsection{CIL through Task-ID Prediction}  
The characteristic of this type of method is the use of additional task-ID prediction to extend TIL methods to CIL.
Early attempts include CCG~\cite{continual2020_4}, which constructs a separate network to predict task-ID, and HyperNet~\cite{hypernet} and PR-Ent~\cite{prent} which use entropy metrics to predict task-ID. The performances of these methods are not satisfactory, and some of these methods also face practical application issues. For example, iTAML~\cite{itaml} requires that the data in the same batch come from the same task during testing, which clearly does not meet the needs of real-world applications. {HILAND~\cite{HILAND} designs a class for ``past'' classes, which is trained using exemplar memory from prior tasks, and determines the task-ID through a hierarchical structure combined with either global threshold optimization and hierarchical voting mechanisms. {However, it introduce error propagation and ambiguous task boundaries due to the lack of explicit modeling of future tasks.}
BEEF~\cite{beef} defines an additional ``forward prototype'' to explicitly model out-of-distribution uncertainty. When an input sample belongs to a future unknown task, this prototype yields high uncertainty scores, thereby reducing the confidence of the current module. As a result, the module designed for that future task can dominate the final prediction.}
MORE~\cite{more} uses a pretrained ImageNet1000~\cite{imagenet1000} model as the base and incorporates task-specific masked adapters into the Transformer blocks, which are trained when new tasks arrive while freezing the main parameters of the model. During testing, it utilizes the softmax output values and corrects them using the Mahalanobis distance between features and class centroids to perform task-ID prediction. {Although computational cost is reduced compared with full fine-tuning, each task still requires a considerable number of parameters (e.g., approximately 24 million trainable parameters plus additional mask parameters).} Additionally, when the number of samples per class is small, the estimated centroid of each class may not be accurate, thereby affecting the effectiveness of score correction.

Similar to existing task-ID prediction-based approaches, our method leverages the concept of OOD detection to facilitate task-ID prediction, enabling the selection of the appropriate task-specific module during inference. However, our method introduces an ``unknown'' class for each task-specific classification head to explicitly represent out-of-task categories. During the alignment stage, classification heads are encouraged to produce consistent ``unknown'' outputs for samples from other tasks, allowing older heads to retain awareness of potential future tasks. Moreover, our approach trains only lightweight task-specific BN layers and classifier heads, resulting in approximately 15K trainable parameters per task - substantially fewer than existing methods and achieving higher efficiency.
\color{black}


In conclusion, this area of research is relatively new as its exploratory theoretical support and performance guarantees are gradually emerging. 
These methods hold a broad research prospect, particularly by enhancing the accuracy of TP during testing, which is expected to further improve performance. However, the similarity between tasks may pose challenges for task-ID prediction, necessitating further in-depth research and resolution.

\section{Method}

\begin{figure}[!th]
    \label{train_test}
    \centering
    \includegraphics[width=1.0\textwidth, keepaspectratio]{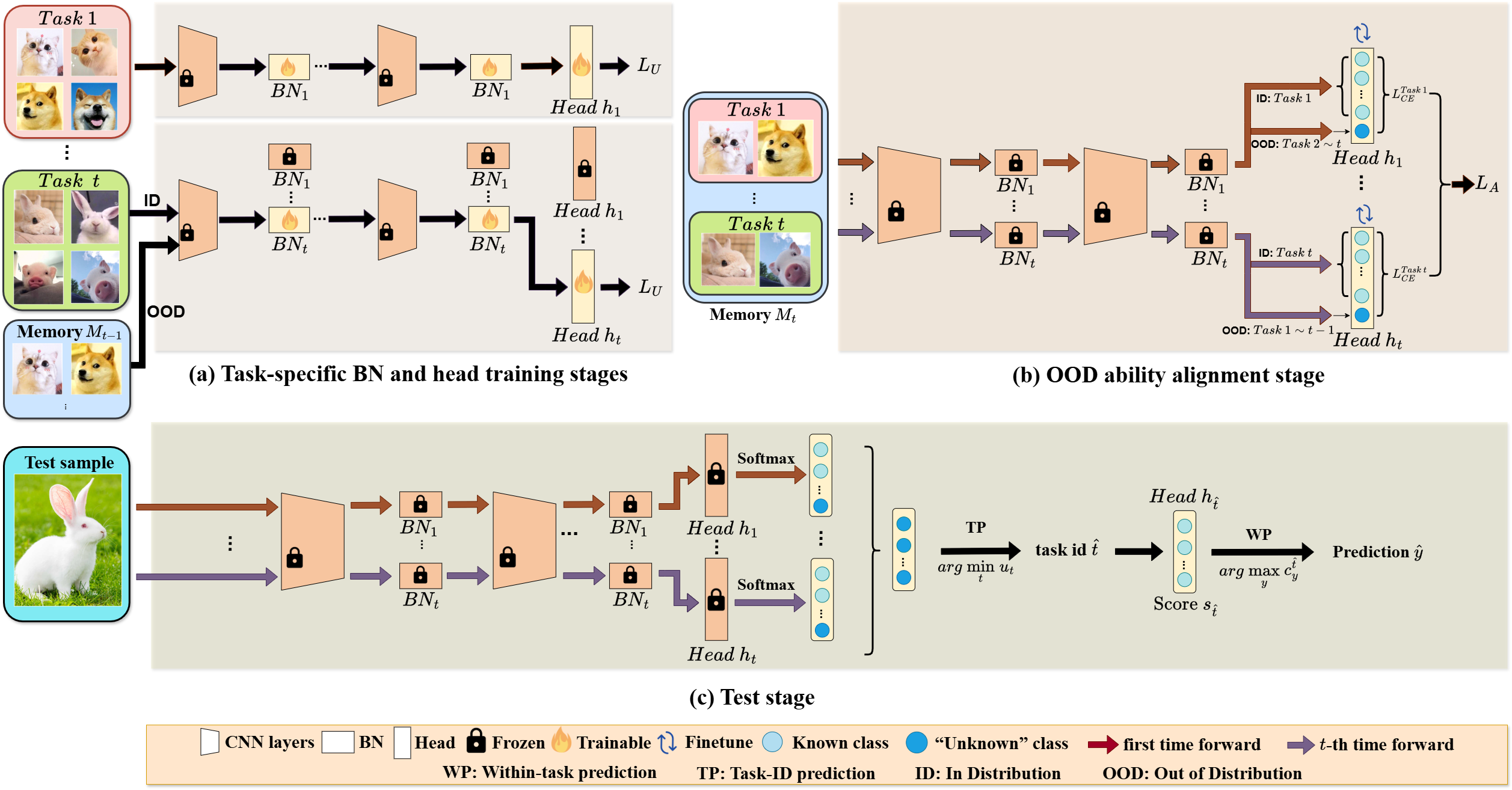}
    \caption{Framework of our proposed method. Sub-figures (a) and (b) depict the training stage. Sub-figure (a) shows the process of gradually adding and training task-specific BN layers and classifier heads as new tasks arrive. After a new task is learned, the OOD capability alignment stage, as shown in sub-figure (b), is conducted. Sub-figure (c) presents the test stage, which starts by identifying the classifier head with the lowest output probability of the ``unknown'' class for task-ID prediction (TP) and finally performing within-task prediction (WP) with the selected classifier head.}
    \label{our_method}
\end{figure}

The schematic diagram of the method is shown in Figure~\ref{our_method}. Figure~\ref{our_method}(a) and Figure~\ref{our_method}(b) represent the training stage. Figure~\ref{our_method}(a) depicts the first stage of training, where task-specific batch normalization and task-specific classifier head are added as a new task arrives. {In particular, an additional ``unknow'' class is included to the task-specific classifier head to represent samples that do not belong to the current task.} Each task corresponds to a task-specific sub-model, where each sub-model consists of task-shared convolutional kernels, task-specific BNs, and a task-specific classifier head. Figure~\ref{our_method}(b) represents the second stage of training, {where a class-balanced subset of samples are drawn from the replay memory to fine-tune all task-specific classifier heads, thereby aligning out-of-distribution detection capabilities across classifier heads.} During the test stage, as illustrated in Figure~\ref{our_method}(c), under the existence of multiple task-specific sub-models, task-ID prediction is performed first, followed by within-task prediction to obtain the final prediction.

\subsection{Task-specific Batch Normalization and Task-specific Head}
The powerful expressive capability of batch normalization was discovered early in research work about generative adversarial networks (GAN) \cite{karras2019style,karras2021alias}. In particular, 
the StyleGAN model is able to generate a variety of realistic images by just merely adjusting the BN layer(s). {Owing to BN’s ability to adjust feature distributions, even BN alone can improve the model’s capacity to learn new tasks, thus ensuring plasticity. Meanwhile, freezing the shared convolutional layers and preserved BN parameters of previous tasks ensures knowledge retention and stability.} Specifically, for the $j$-th convolutional kernel of the $l$-th layer within the convolutional neural network, we denote the output of the convolution operation on a single input sample as feature map $\mathbf{z}_{l, j}$. Across a batch, the outputs of all input samples for this kernel are marked as $\mathbf{Z}_{l, j}$. The BN layer standardizes each feature map as follows
\begin{equation}\label{eq:bn}
q_{l,j}(\mathbf{z}_{l, j}) = \gamma_{l,j} \frac{\mathbf{z}_{l, j} - \mu(\mathbf{Z}_{l, j})}{\sigma(\mathbf{Z}_{l, j})} + \beta_{l,j},
\end{equation}
where $\mu(\mathbf{Z}_{l, j})$ and $\sigma(\mathbf{Z}_{l, j})$ denote the mean and standard deviation of all elements within the feature map produced by the $j$-th convolutional kernel of the $l$-th layer. $\gamma_{l,j}$ and $\beta_{l,j}$ are trainable scaling and shifting parameters that independently adjust each feature map. This means that BN first normalizes the outputs to zero mean and unit variance and then re-adjusts the distribution of the feature maps through scaling and shifting operations. This formula (Equation~\ref{eq:bn}) shows that the parameters of the BN layer(s) are directly related to the data distribution, enhancing the adaptability of the feature extractor for different task data distributions. 

Theoretically, given that the objective function of classification task is usually set as the cross-entropy loss between softmax output vector and the one-hot ground-truth vector, task-specific BN parameters will be optimized distinctively for each task-specific local optimal point of the cross-entropy loss to fit the task-specific data distribution. By contrast with task-shared BN (regular CNN backbones), the local optimal point of the joint cross-entropy loss of all tasks is likely not close to each task-specific local optimal point, particularly considering that most data of old tasks are not available during learning a new task. This is intuitively indicated when the data distributions across tasks are various, where shared-BN cannot fit well but task-specific BN is able to fit perfectly each task data distribution.

Moreover, compared to 
convolutional kernels, the BN layers contain much fewer parameters. This makes BN layers not only easier to be adjusted but also more storage-efficient. For example, in the ResNet18 model, the total parameter number of all convolutional kernels is about 11 millions, but all the BN layers contain only about 15,000 parameters, much fewer than that of all convolutional kernels.
 
Considering the benefits of BN layers in improving expressiveness and facilitating parameter control, we propose that, 
after obtaining a strong feature extractor through pretraining, 
task-specific BN layers and task-specific classifier head are added and trained for each new task. 
Due to the small number of new learnable parameters, these learned task-specific BN parameters and the classifier head can be stored throughout all incremental learning stages. This strategy is not only efficient in terms of storage but also enhances the learning capability of the feature extractor for each task through the robust expressive power of the BN layers. Since task-specific BN layers and task-specific head are stored and not changed for each old task when the model learns a new task, knowledge of old tasks is well preserved and not forgotten in the incremental learning process.

The following provides a symbolic representation of how the proposed method augments the relevant elements of the model as a new task arrives in the incremental learning framework. In incremental learning, data sequentially arrive in $T$ tasks, with each task associated with a specific dataset, represented as ${\mathbf{D}_1, \mathbf{D}_2, \ldots, \mathbf{D}_T}$. In the feature extractor, the convolutional kernels, primarily responsible for extracting image features, are parameterized by 
\(\bm{\phi}\). The batch normalization, responsible for standardizing the output of convolutions, is parameterized by 
\(\bm{\omega}_t\). With the arrival of the new $t$-th task, the convolutional kernel parameters \(\bm{\phi}\) remain fixed and do not participate in training. The introduction of a new task results in the addition and training of BN layer parameters \(\bm{\omega}_t\) and the corresponding classifier head $h_t$. 
Therefore, after completing the training for the $t$-th task, the feature extractor will include one 
\(\bm{\phi}\) and $t$ sets of task-specific BN parameters $\{\bm{\omega}_k, k=1, 2, \ldots, t\}$. 
By combining \(\bm{\phi}\) with each \(\bm{\omega}_k\) and \(h_k\), the model effectively forms \(t\) distinct sub-models, with each sub-model specialized for a specific task. The $k$-th sub-model, denoted as \(S_k\), is thus responsible for handling the $k$-th task, utilizing the corresponding batch normalization parameters and the classifier head.

\subsection{``Unknown'' Class in Task-specific Classifer Head}
\label{sec33}
Our method employs multiple task-specific classifier heads for class incremental learning. Due to the unknown task-ID in test stage, it is necessary to first perform task-ID Prediction (TP) to select a task head, and then proceed with within-task Prediction (WP) to achieve the final prediction. 
The test stage is illustrated in Figure~\ref{our_method}(c). The accuracy of TP directly affects whether the task that contains ground-truth class of the test sample is selected, and an incorrect selection renders WP ineffective. Moreover, the accuracy of WP indicates the ability to accurately classify within the task once the appropriate task head has been correctly chosen. 
Thus, the effectiveness of the proposed method fundamentally depends on the multiplication of WP accuracy and TP accuracy. 
Considering that the performance of WP can be ensured with task-specific BN and classifier head, 
TP becomes crucial for the proposed method. 
To make TP perform well, each task-specific sub-model \(S_k\) for the $k$-th task must have the capability to detect out-of-distribution (OOD) samples. This capability enables the model to effectively identify whether a new sample is from certain class of the current $k$-th task or from other tasks.

To equip each sub-model with OOD detection capability, we incorporate the concept of the ``unknown'' class from the classical OpenMax~\cite{openmax} method into our model architecture. Similar to OpenMax, the task-specific classifier head is expanded by adding an output unit for a special ``unknown'' class, where the ``unknown'' class represents the class of all OOD samples for the $t$-th task.
In particular, samples from all previously learned $t-1$ tasks are considered OOD samples for the current $t$-th task, and vice versa. Considering a small number of old task samples stored in memory are available when learning the new $t$-th task, all the stored old samples can be naturally adopted as training samples of the ``unknown'' class for the $t$-th task.
{Note that even there is no data for the ``unknow'' class of the first task, the first task classifier head can still be trained normally. The ``unknow'' class becomes effective in all of the subsequent tasks during the OOD Detection Alignment Stage, for a given task task-specific classifier head, samples from any other tasks are mapped to the ``unknow'' class as the groundtruth during alignment finetuning of that task classifier.}
Formally, let \(\mathbf{M}_{t-1}\) denote the memory set that contains a small number of samples from the previous \(t-1\)  tasks, and let \(\mathbf{D}_t\) denote the set of samples for the new task that contains \(C_t\) classes. In the new classifier head \(h_t\) for task \(t\), the \((C_{t}+1)\)-th class is introduced as the ``unknown'' class. The learnable parameters for task \(t\) are \(\bm{\theta}_t = \{\bm{\omega}_t, h_t\}\), and the corresponding loss function is designed as
\begin{equation}
\begin{aligned}
L_\text{U}(\bm{\theta}_t) = & -\frac{1}{|\mathbf{D}_{t}|}\sum_{(\mathbf{x}_i, y_i) \in \mathbf{D}_{t}} \log p(y_i | \mathbf{x}_i; \bm{\phi}, \bm{\omega}_t, h_t) \\
                         & -\frac{1}{|\mathbf{M}_{t-1}|}\sum_{\mathbf{x}_j \in \mathbf{M}_{t-1} } \log p(C_t+1 | \mathbf{x}_j;\bm{\phi}, \bm{\omega}_t, h_t),
\end{aligned}
\end{equation}
where \(\mathbf{x}_i\) and \(y_i\) represent $i$-th sample and its label from task \(t\), while \(\mathbf{x}_j\) represents the $j$-th old sample from the memory $\ve{M}_{t-1}$. With the help of stored old task samples as OOD training data, the well-trained task-specific sub-model $S_t$ will be able to effectively distinguish whether a new sample is from the current $t$-th task or from previously learned tasks. Consequently, the outputs from the $t$ ``unknown'' classes of all the learned $t$ classifier heads can be used to perform TP.
Specifically, during the test stage, after a new data is passed through the sub-models of all tasks, the task head with the smallest output of the ``unknown'' class over all $t$ ``unknown'' classes is first selected (TP). Then, within-task prediction (WP) is performed with the selected task-specific classifier head to obtain the final class prediction.

\subsection{OOD Detection Alignment Stage}
\label{ood-alignment}

During training of the new task-specific BN and classifier head, samples from old tasks are mapped toward the ``Unknown'' class in the new classifier head. Although the memory can only store a limited number of old task samples for replay, the number of old task classes in the memory gradually increases with more tasks learned. 
This means that sub-models specific to later tasks encounter a broader range of old class samples. For instance, in the sub-model corresponding to the first task, there are no old task samples in the memory, while the sub-model for the final task has access to stored samples from all previous task classes. This imbalance can lead to variations in the ``Unknown'' class output from different task-specific classifier heads, resulting in inconsistent confidence outputs and a decrease in the accuracy of task identification prediction. To address this challenge, an OOD detection alignment stage is introduced to fine-tune each task-specific classifier head.

Specifically, this stage is added after training the BN layers and the classifier head for the new task, as shown in Figure~\ref{our_method}(b). During this stage, the feature extractor remains fixed, and only the parameters of the classifier heads are updated.
Taking the $t$-th task as an example, the herding selection~\cite{icarl} iteratively selects samples from the $t$-th task data that are closest to the class mean in the frozen feature space to update \(\ve{M}_{t-1}\) to \(\ve{M}_t\). Denote the newly selected sample set for the $t$-th task in the memory as $\mathbf{R}_t$, hence $\displaystyle \mathbf{M}_t=\bigcup_{k=1}^t\mathbf{R}_k = \mathbf{M}_{t-1}\cup \mathbf{R}_t$.  This ensures balance coverage of all learned \(t\) tasks within the memory budget. 
Subsequently, \(\ve{M}_t\) is used to train all $t$ classifier heads through the OOD alignment replay by minimizing the loss function $L_\text{A}(\bm{\theta}_t)$, i.e.,  
\begin{align}\label{eq:align}
\small
L_\text{A}(\bm{\theta}_t)= & -\frac{1}{t \cdot |\mathbf{M}_t|} \sum_{k=1}^t \Bigg[  \sum_{(\mathbf{x}_i, y_i) \in \mathbf{R}_k} 
 \log p\left(y_i \mid \mathbf{x}_i; \bm{\phi}, \bm{\omega}_k, h_{k}\right) \notag \\
 & +\sum_{(\mathbf{x}_j, y_j) \in (\mathbf{M}_t \backslash \mathbf{R}_k) } \log p\left(C_k+1 \mid \mathbf{x}_j; \bm{\phi}, \bm{\omega}_k, h_{k}\right) \Bigg] \,,
\end{align}
where $\mathbf{M}_t \backslash \mathbf{R}_k$ represents the memory excluding the stored samples from task $k$. For each task-specific classifier head \(h_k\), 
stored training samples from all the other $t-1$ tasks are associated with the ``unknown'' class, denoted as class \(C_{k}+1\), in the classifier head \(h_k\). Therefore, the second loss term in Equation~(\ref{eq:align}) can help the task-specific classifier head $h_k$ more accurately identify whether a new sample is from one class of the current task $k$ or from any other task. Since all the $t$ classifier heads are fine-tuned simultaneously and the number of training samples for the ``unknown'' class is equivalent for all the $t$ task-specific classifier heads, the outputs from the $t$  ``unknown'' classes across the $t$ classifier heads are aligned and can be directly compared with each other during task-ID prediction. When one sample is from task $k$, the output of the ``unknown'' class from the $k$-th classifier head will probably be lower (i.e., close to zero), while the output of the ``unknown'' class from any other classifier head will probably be much higher, thus improving the performance of task-ID prediction.

\subsection{Comparison with Related Methods}
{Our method introduces key innovations in both the use of task-specific BN and task-ID prediction, setting it apart from existing approaches.}

{Improving BN layers for task-specific data distributions in continual learning has been widely studied, with methods such as NBN~\cite{NBN}, AdaB2N~\cite{AdaB2N}, TBBN~\cite{TBBN}, and CN~\cite{CN} addressing BN-induced forgetting through noise injection, Bayesian weighting, virtual batches, or hybrid normalization, but all rely on task-shared BN layers, limiting their ability to capture diverse feature distributions. By contrast, our method introduces task-specific BN parameters that adaptively model the feature distributions of object classes within each task, enabling more discriminative within-task predictions. Unlike TSN-IQA~\cite{TSN-IQA}, which uses task-specific BN for feature distributions of particular distortion types modeling with a regression-based objective and a gated ensemble inference strategy, our approach employs a two-stage training process: (1) training task-specific BN and classifier heads with cross-entropy loss and an additional ``unknow'' class for OOD detection, and (2) OOD detection alignment using balanced replay to calibrate classifier heads, thereby improving both within-task accuracy and task-ID reliability for inference.}

{On the other hand, extending TIL methods to CIL via task-ID prediction has been explored in works such as Kim et al.~\cite{cil_ood_theory}, which combines the TIL method HAT~\cite{hat} with the contrastive learning method CSI~\cite{csi} and achieves favorable results, MORE~\cite{more}, which adds task-specific masked adapters to ViT~\cite{vit} and leverages the HAT approach to train masks that protect important parameters by masking their gradients, preventing updates, and HILAND~\cite{HILAND}, which employs hierarchical sequencing to construct multiple incremental CNN classifiers for task inference during testing. Compared to MORE, our method only trains task-specific BNs and classifiers for each task, requiring merely about 15,000 trainable parameters, which is substantially fewer than the 24 million gradient-computed parameters and 70,000 additional mask parameters required by MORE. Despite this drastic reduction, our method achieves superior performance across multiple datasets. Moreover, unlike HILAND, which distinguishes tasks by outputting a generic “past tasks” class without explicit alignment to future task data, and relies on threshold optimization and hierarchical ranking, thereby lacking fine-grained modeling of task-specific distribution shifts and causing blurred task boundaries as well as error propagation, our method employs task-specific BNs to model the feature distributions of classes within each task and equips classifier heads to jointly predict within-task and out-of-task classes through an ``unknow'' class. An additional alignment stage enforces consistent outputs across tasks, thereby enhancing both interpretability and task-ID reliability.}

\section{Experiments}
\subsection{Experiment setup}

\noindent \textit{Datasets}: We conducted a series of experiments on two medical image datasets, including Skin8~\cite{tschandl2018ham10000} and Path16~\cite{yangyang,borkowski2019lung,janowczyk2016deep,wei2021petri,veeling2018rotation, bejnordi2017diagnostic,Oral_Cancer}, as well as two natural image datasets, CIFAR100~\cite{cifar} and {CUB200}~\cite{cub200}, to thoroughly validate the effectiveness of our proposed method. 
These four datasets are all publicly available on the Internet, and their statistics are presented in Table~\ref{dataset-sta}.

\begin{table}[h]
\centering
\caption{Statistics of four datasets: Skin8, Path16, CIFAR100 and {CUB200}. [600, 1024] and {[}120, 500{]} represents the range of image dimensions in terms of height and width.}

\label{dataset-sta}
    \resizebox{0.7\linewidth}{!}{

        \begin{tabular}{ccccc}
        \toprule
Dataset & Number of Classes & Train Set & Test Set & Image Size \\
        \midrule

        Skin8  & 8   & 3,555  & 705 & {[}600, 1024{]}   \\
        Path16 & 16   & 11,449  & 1,607 & $224\times224$ \\
        CIFAR100 & 100   & 50,000  & 10,000 & $32\times32$ \\
        {CUB200} & 200   & 5,994  & 5,794 &   {[}120, 500{]} \\
        \bottomrule
        \end{tabular}
    }
        
\end{table}

Skin8~\cite{tschandl2018ham10000} is the dataset used in the 2019 skin disease classification challenge hosted by the International Skin Imaging Collaboration. It is a large dataset of skin lesion images sourced from various locations worldwide. 

Path16~\cite{yangyang,borkowski2019lung,janowczyk2016deep,wei2021petri,veeling2018rotation, bejnordi2017diagnostic,Oral_Cancer} derives from various public pathology datasets, covering a range of disease types and diagnostic scenarios. The dataset statistics are shown in Table~\ref{tab:med_dataset_path16}. It includes 7 sub-tasks, each simulating the diagnostic needs for specific diseases in different body parts, such as the stomach and lungs, with high-resolution medical images except Breast and Lymph Node sub-tasks. 
\begin{table}[h]
    \centering
    \caption{Statistics of the Path16 Dataset. {[}224, 2048{]} represents the range of image dimensions in terms of height and width.}
    \resizebox{\linewidth}{!}{
    \begin{tabular}{ccccc}
        \toprule  
        Task Name & Number of Classes & Image Size & Train Set & Test Set\\ 
        \midrule  
        Stomach (ST)\cite{yangyang} & 4 & 224×224 & 3,208 & 407 \\
        Colon (CO)\cite{borkowski2019lung} & 2 & 768×768 & 1,600 & 200 \\
        Lung (LU)\cite{borkowski2019lung} & 2 & 768×768 & 1,600 & 200 \\
        Breast (BR)\cite{janowczyk2016deep} & 2 & 50×50 & 1,600 & 200 \\
        Colorectal Polyps (CP)\cite{wei2021petri} & 2 & 224×224 & 1,000 & 200 \\
        Lymph Node (LN)\cite{veeling2018rotation,bejnordi2017diagnostic} & 2 & 96×96 & 800 &200 \\
        Oral Cavity (OC)\cite{Oral_Cancer} & 2 & {[}224, 2048{]} & 1,641 & 200 \\
    \bottomrule  
    \end{tabular}
    }
    \label{tab:med_dataset_path16}
\end{table}

CIFAR100~\cite{cifar} consists of 60,000 color images at a resolution of 32x32 pixels, distributed across 100 different classes. Each class contains 600 images, with 500 used for training and 100 reserved for testing. The dataset is diverse, containing a wide variety of images, ranging from animals to indoor and outdoor objects in natural scenes.

{The CUB200 dataset~\cite{cub200} is a fine-grained image benchmark containing 11,788 images from 200 North American bird species. Each image contains a single bird instance, while exhibiting significant variations in background, pose, and illumination, making it a high-quality yet challenging benchmark for fine-grained visual recognition.}

\noindent \textit{Hyperparameter Settings}:  
For the Path16 and Skin8 datasets, the initial training stage consists of 200 epochs with a batch size of 32. For the optimizer, we utilize Stochastic Gradient Descent~\cite{sgd} (SGD) with an initial learning rate of 0.01, momentum of 0.9, and weight decay of 0.0005. The learning rate is adjusted according to a multi-step scheduler, dropping to one-tenth of its original value at epochs 70, 130, and 170. The confidence alignment stage lasts for 100 epochs, starting with a learning rate of 0.001 and decreasing to one-tenth at the $55^{th}$ and $80^{th}$ epochs. For the CIFAR100 and {CUB200} dataset, while maintaining the same optimizer settings and epoch durations in the initial training stage, the batch size is changed to 128 and the weight decay is changed to 0.0002. The OOD detection alignment stage lasts for 50 epochs, with learning rate reductions at the $15^{th}$ and $35^{th}$ epochs.

\noindent \textit{Data Augmentation Strategies}: 
The data preprocessing and augmentation applied to the Path16 dataset includes 50\% random horizontal flips, resizing images to 224 $\times$ 224, and normalization. 
The preprocessing on Skin8 dataset mirrors these techniques but we adjust the mean and standard deviation values for normalization to cater to its unique image characteristics. The CIFAR100 and {CUB200} dataset introduces a series of augmentations including random cropping, brightness adjustments, and resizing. 

\noindent \textit{Metrics}: 
To address class imbalance in the Skin8 dataset, we adopt Classification Accuracy (ACC) as our primary evaluation metric. 
After completion of incremental learning of all $T$ tasks, the ACC over all learned classes of the $T$ tasks on the whole test set, namely Last-ACC, is adopted as one metric for learning performance. 
In addition, the ACC after learning each task is obtained and then the average of these ACCs over all $T$ tasks, namely Avg-ACC, is also used to measure incremental learning performance of the proposed method. 

To evaluate the performance of task-ID prediction (TP), the task-ID prediction accuracy $\beta_t$ is used for task $t$, measuring the performance of the model in correctly identifying test samples of task $t$ among all actual test samples of task $t$. Similarly as Last-ACC and Avg-ACC, we define Last-TP as $\frac{1}{T} \sum_{k=1}^T \beta_k$ and Avg-TP as $\frac{1}{T} \sum_{t=1}^T \{\frac{1}{t}\sum_{k=1}^t \beta_k\}$.

\subsection{Effectiveness Analysis}

Our proposed method was compared with two types of methods for the CIL task scenario and setting. The first type of CIL methods leverages a task-ID prediction method to apply multi-head TIL methods to the CIL domain, including MORE~\cite{more}, HILAND~\cite{HILAND}, and BEEF~\cite{beef}. The other type of CIL methods directly perform an incremental classification task with a single-head model architecture, including iCaRL~\cite{icarl}, DynaER~\cite{DER}, WA~\cite{wa}, DER++~\cite{darker}, UCIR~\cite{ucir}, PODNet~\cite{podnet}, Dytox~\cite{dytox}, DRC~\cite{drc}, DSGD~\cite{dsgd}, CREATE~\cite{create}, and VBM-TCIL~\cite{vbm}.

To ensure fair comparison, we unified the backbone architecture across methods while preserving their original characteristics:
\begin{enumerate}
    \item For MORE~\cite{more}, we followed its original implementation using DeiT-S~\cite{deits} with task-specific masked adapters.
    \item All other baseline methods (iCaRL, DynaER, etc.) retained ResNet18~\cite{resnet} backbones pretrained on ImageNet-1K~\cite{imagenet1000}, matching our method's configuration.
\end{enumerate}

\begin{figure}[!ht]
    \centering
    \subfloat[CIFAR100 dataset, 10 tasks.\label{cifar100-10t}]{
        \includegraphics[width=0.45\linewidth, height= 3.3cm]{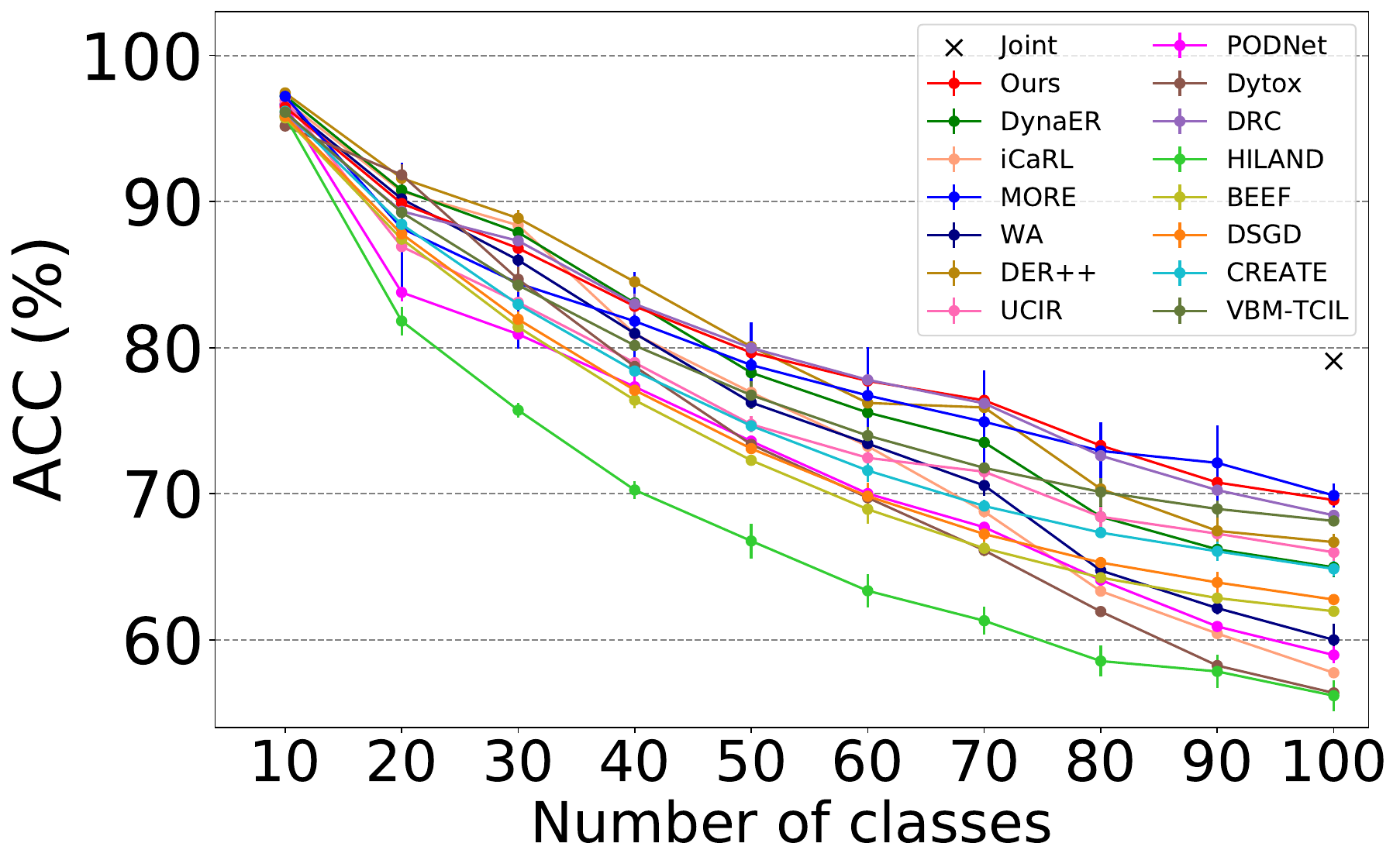}
    }
    \hfil
    \subfloat[CIFAR100 dataset, 20 tasks.\label{cifar100-20t}]{
        \includegraphics[width=0.45\linewidth, height= 3.3cm]{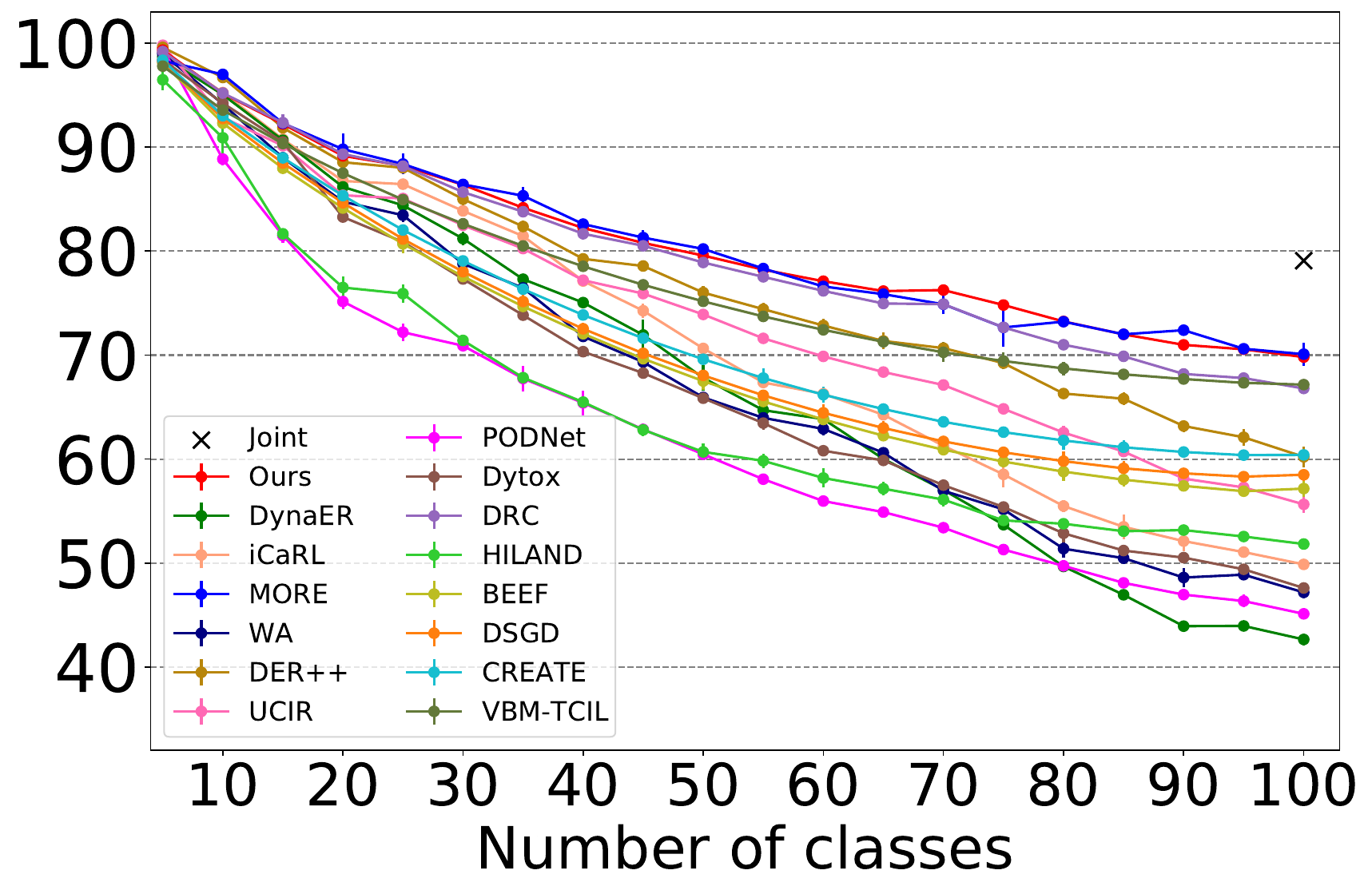}
    }
    \hfil
    \subfloat[CUB200 dataset, 10 tasks.\label{cub200-10t}]{
        \includegraphics[width=0.45\linewidth, height= 3.3cm]{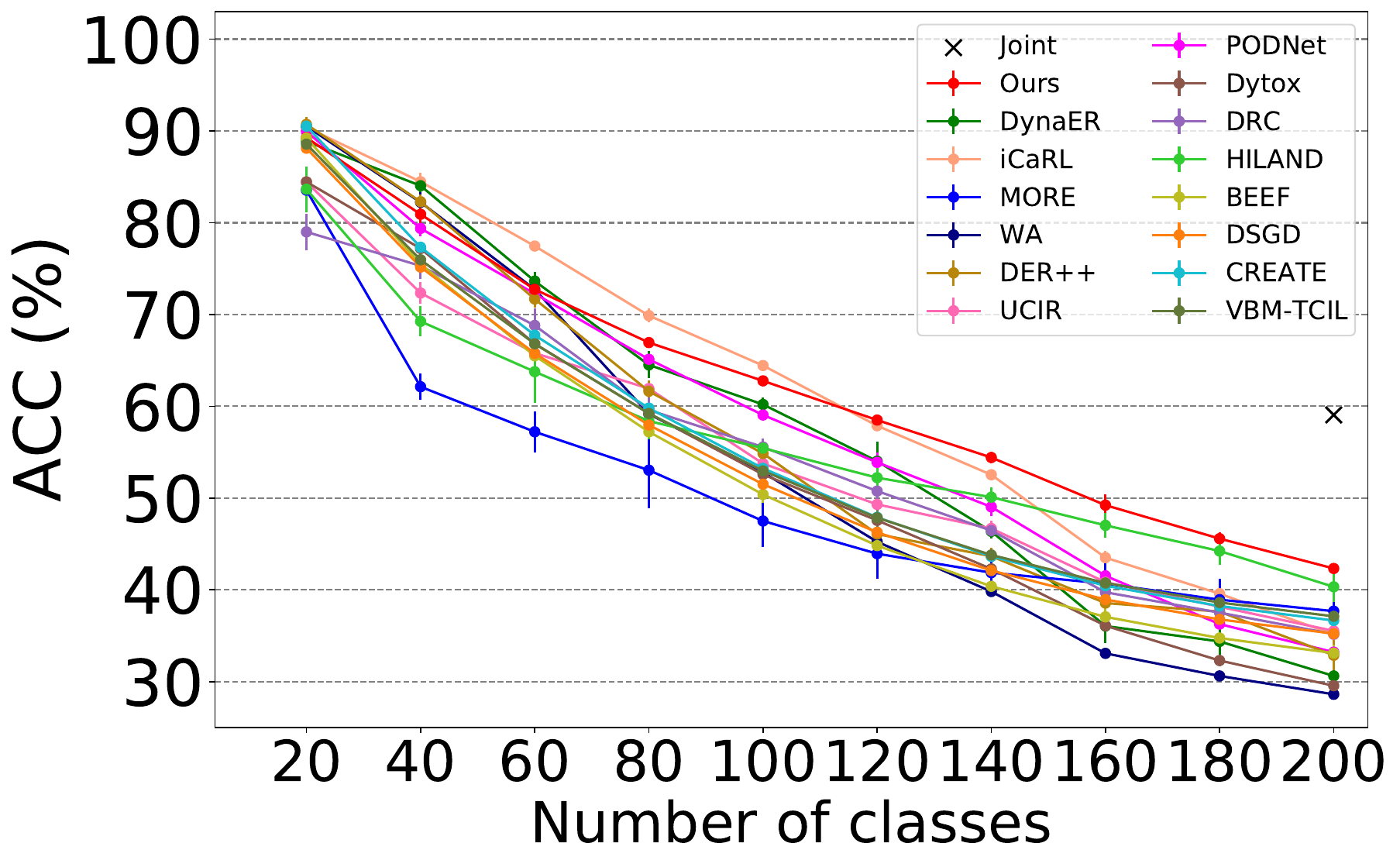}
    }
    \hfil
    \subfloat[CUB200 dataset, 20 tasks.\label{cub200-20t}]{
        \includegraphics[width=0.45\linewidth, height= 3.3cm]{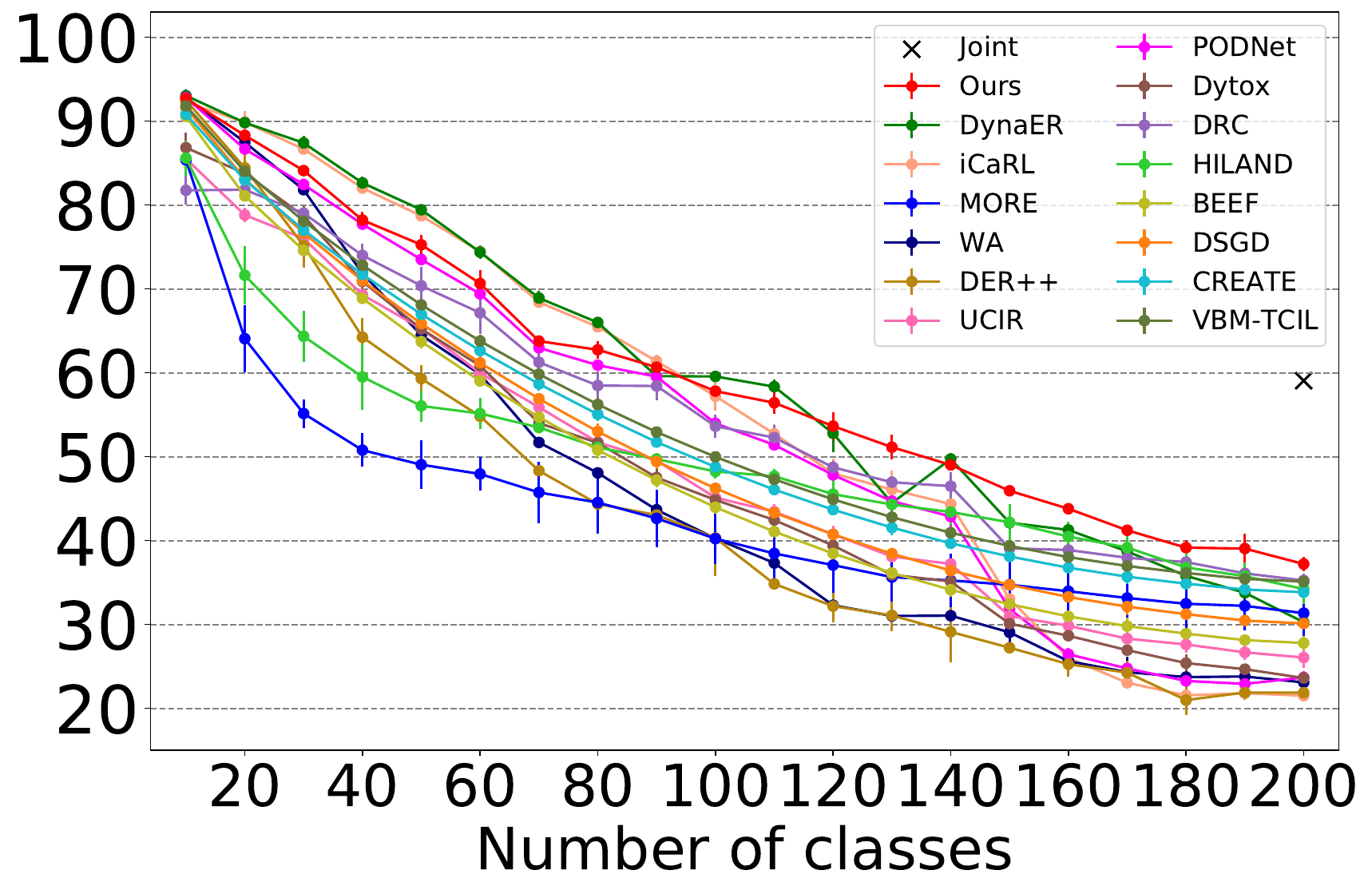}
    }
    \hfil
    \subfloat[Path16 dataset, order I.\label{path16-I}]{
        \includegraphics[width=0.45\linewidth, height= 3.3cm]{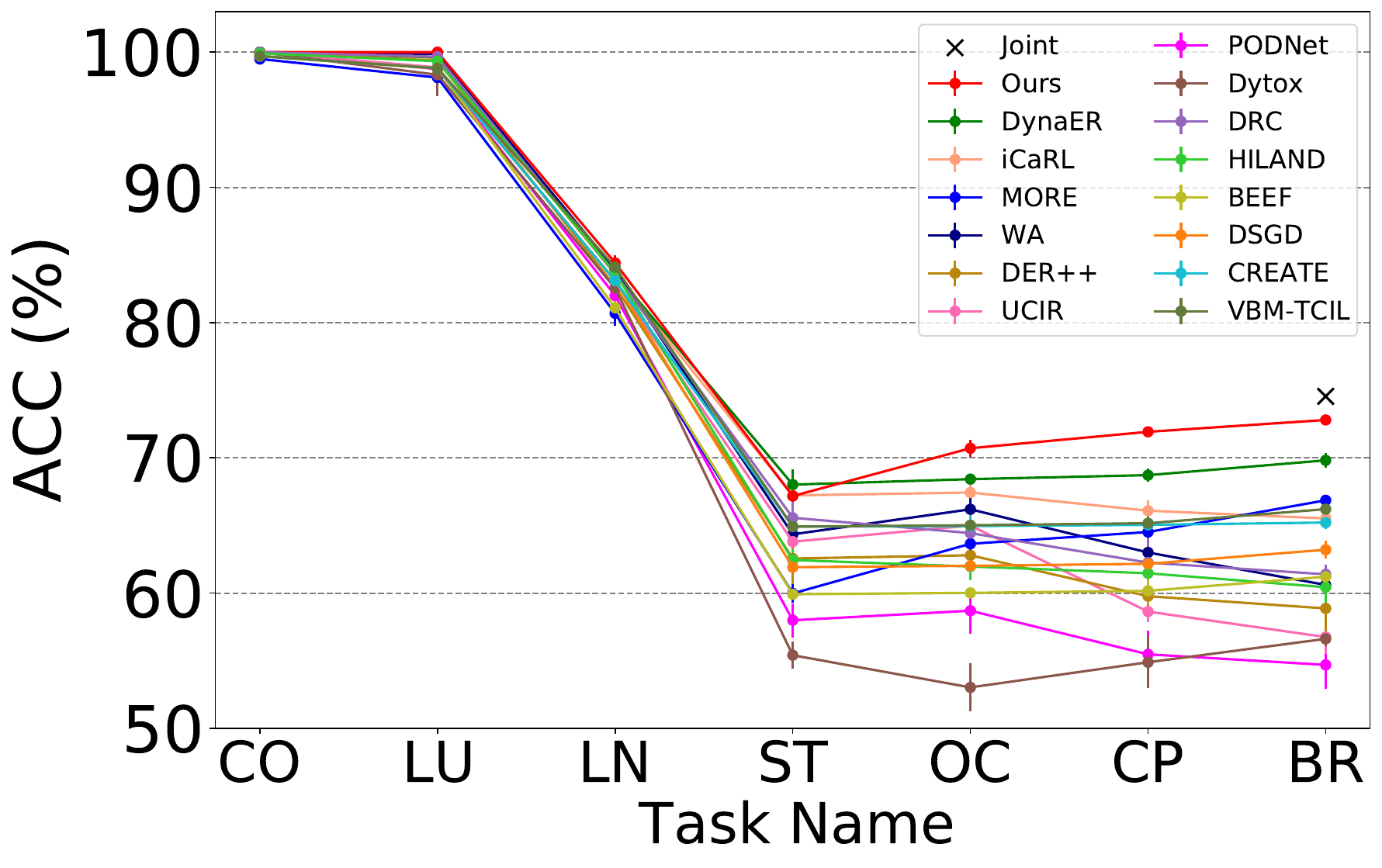}
    }
    \hfil
    \subfloat[Path16 dataset, order II.\label{path16-II}]{
        \includegraphics[width=0.45\linewidth, height= 3.3cm]{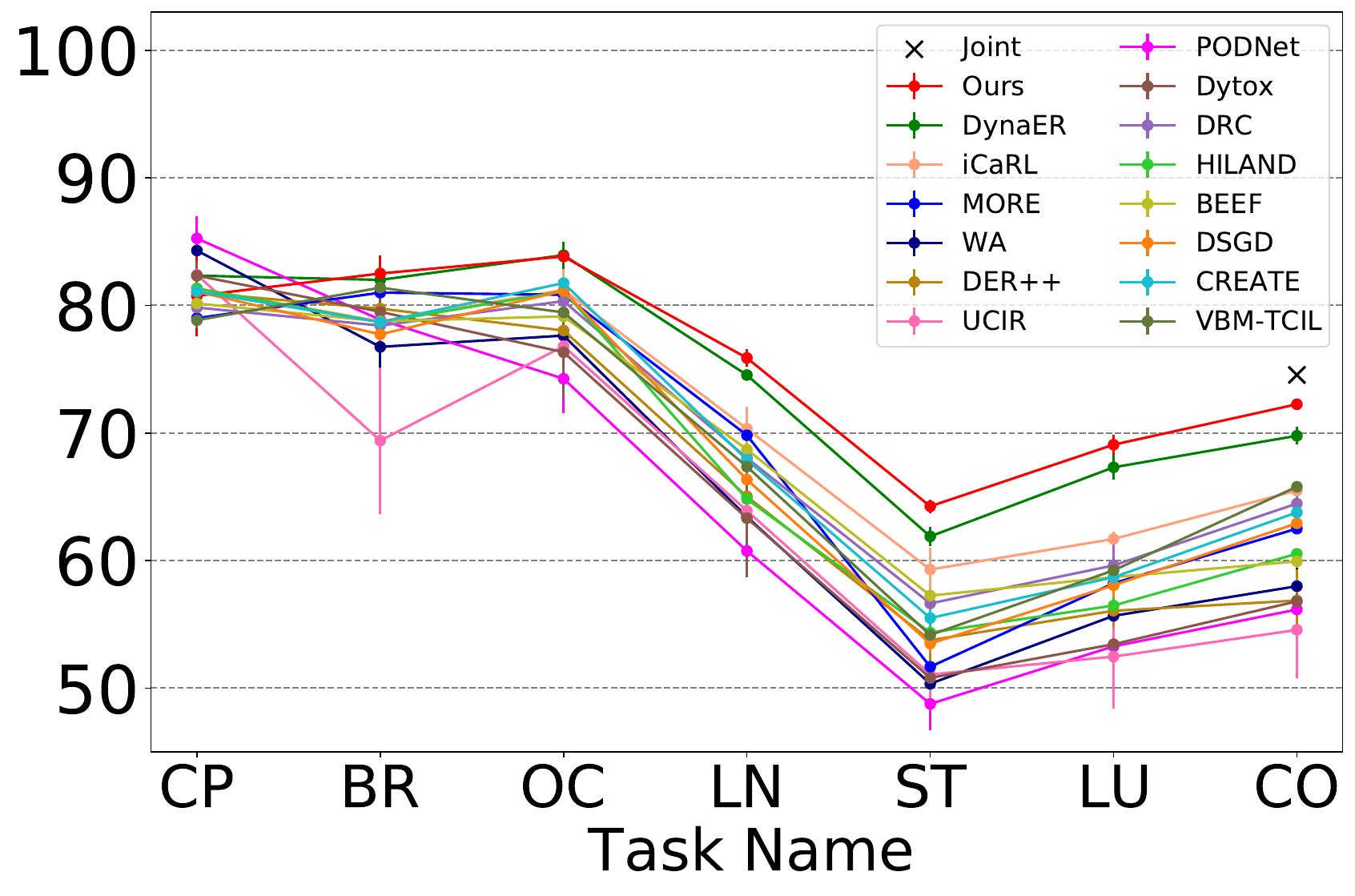}
    }
    \hfil
    \subfloat[Skin8 dataset, memory size 40.\label{skin8-m40}]{
        \includegraphics[width=0.45\linewidth, height= 3.3cm]{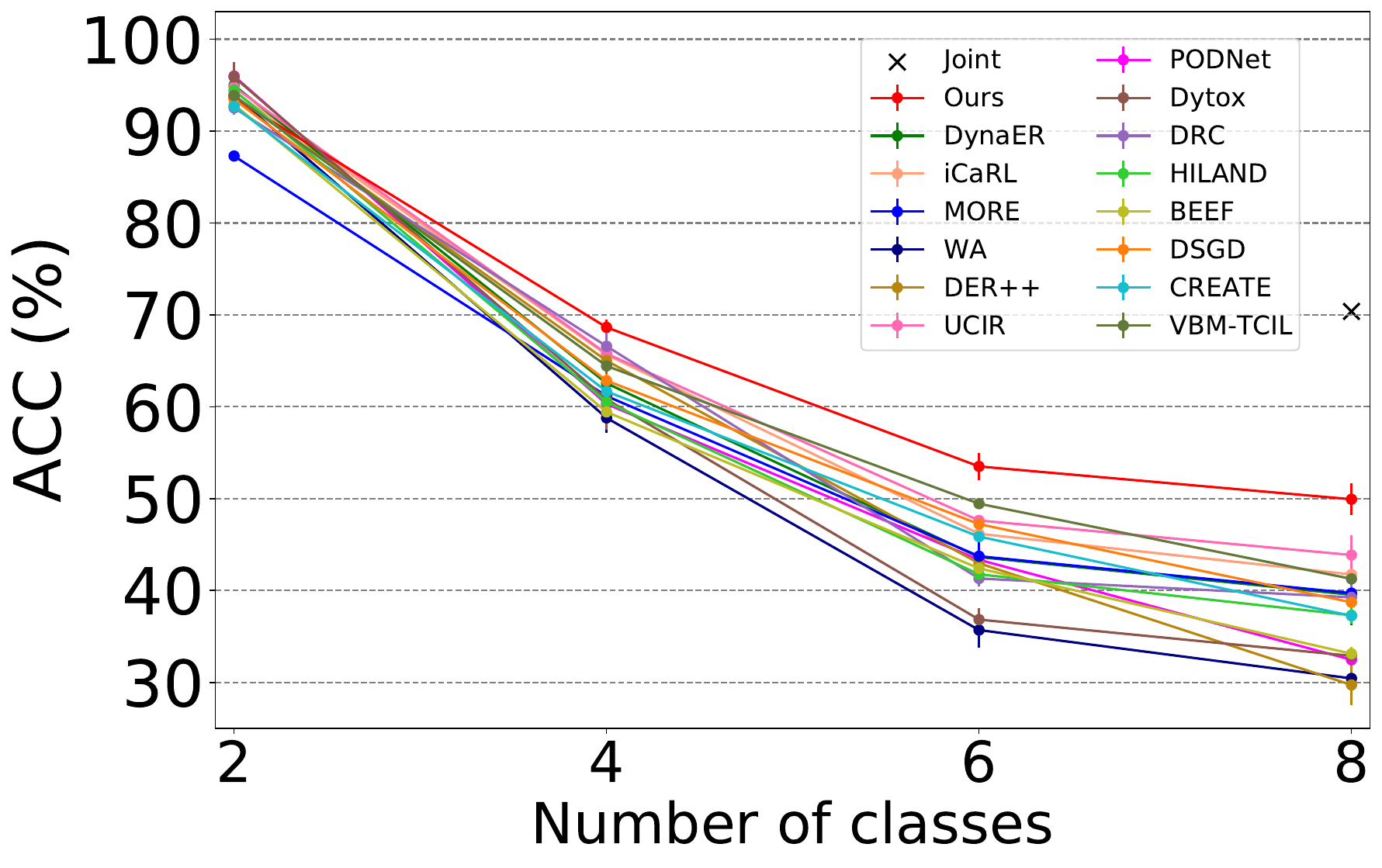}
    }
    \hfil
    \subfloat[Skin8 dataset, memory size 16.\label{skin8-m16}]{
        \includegraphics[width=0.45\linewidth, height= 3.3cm]{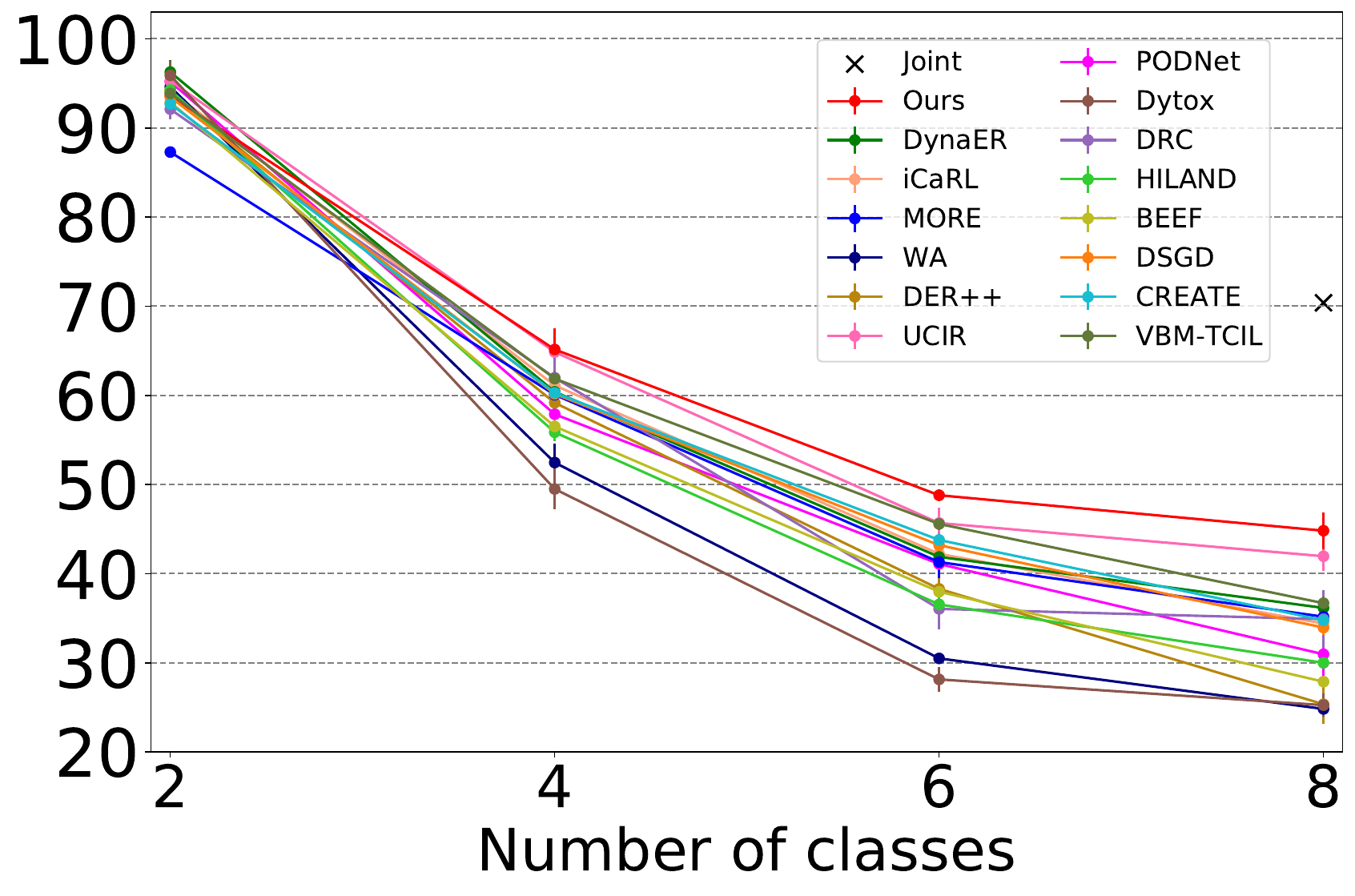}
    }
    \caption{Performance variation curves of different methods on four datasets under different settings.}
    \label{eff-fig}
\end{figure}

Memory configurations were standardized across datasets to ensure fair comparison: For the Path16 dataset, we allocated 80 samples (5 per class across 16 classes) for exemplar storage; the Skin8 dataset used two configurations with 40 samples (5 per class) and 16 samples (2 per class) respectively to evaluate performance under various memory constraints; while the CIFAR100 benchmark employed 2,000 samples (20 per class for its 100 classes) {and the CUB200 benchmark employed 400 samples (2 per class for its 200 classes).}
Results in Figure~\ref{eff-fig}, Table~\ref{eff-tab-path16-skin8} and Table~\ref{eff-tab-cifar100-cub200} are averaged over three seeds. We also report joint training (training on all tasks simultaneously) as the theoretical upper bound (marked ``×'').

\begin{table}[!b]
\centering
\caption{Performance comparison on CIFAR100 and CUB200 dataset. ``CIFAR100-XT'' and
``CUB200-XT'' indicates the CIFAR100 or CUB200 dataset is divided into X tasks.
``Last'' and ``Avg'' respectively represent Last-ACC and Avg-ACC. 
}

\label{eff-tab-cifar100-cub200}
\resizebox{0.9\linewidth}{!}{
\begin{tabular}{ccccccccc}
\toprule
\multirow{2}{*}{Method}
& \multicolumn{2}{c}{CIFAR100-10T} & \multicolumn{2}{c}{CIFAR100-20T}& \multicolumn{2}{c}{CUB200-10T} & \multicolumn{2}{c}{CUB200-20T}  \\
& Last & Avg & Last & Avg & Last & Avg & Last & Avg \\
\midrule
iCaRL 
            & $57.75_{\pm0.35}$ & $75.74_{\pm0.18}$ 
            & $49.87_{\pm0.34}$ & $71.27_{\pm0.42}$

            & $35.15_{\pm0.19}$ & $\underline{61.53_{\pm0.21}}$
            & $21.51_{\pm0.64}$ & $54.73_{\pm0.55}$ \\
UCIR 
            & $65.99_{\pm0.60}$ & $76.63_{\pm0.41}$ 
            & $55.65_{\pm0.81}$ & $73.95_{\pm0.12}$

            & $35.53_{\pm0.28}$ & $54.88_{\pm0.35}$
            & $26.06_{\pm1.23}$ & $48.31_{\pm0.44}$ \\ 
PODNet 
            & $58.96_{\pm0.58}$ & $73.39_{\pm0.32}$ 
            & $45.12_{\pm0.50}$ & $62.72_{\pm0.09}$

            & $33.22_{\pm0.67}$ & $57.97_{\pm0.62}$
            & $23.68_{\pm0.78}$ & $52.99_{\pm0.25}$\\ 
DynaER 
            & $64.97_{\pm0.68}$ & $78.59_{\pm0.23}$ 
            & $42.65_{\pm0.58}$ & $67.74_{\pm0.31}$

            & $30.62_{\pm0.96}$ & $57.27_{\pm0.70}$
            & $30.27_{\pm0.62}$ & $\underline{59.41_{\pm0.17}}$ \\ 
WA 
            & $60.00_{\pm1.11}$ & $76.10_{\pm0.52}$ 
            & $47.18_{\pm0.61}$ & $67.94_{\pm0.39}$ 

            & $28.62_{\pm0.47}$ & $53.48_{\pm0.25}$
            & $23.10_{\pm0.74}$ & $46.17_{\pm0.50}$ \\ 
DER++ 
            & $66.69_{\pm0.54}$ & $79.91_{\pm0.36}$ 
            & $60.21_{\pm0.98}$ & $77.10_{\pm0.10}$

            & $32.87_{\pm1.63}$ & $56.00_{\pm0.85}$
            & $21.87_{\pm0.99}$ & $43.76_{\pm1.40}$ \\ 
MORE 
            & $\mathbf{69.89_{\pm0.80}}$ & $79.70_{\pm2.71}$ 
            & $\mathbf{70.09_{\pm1.12}}$ & $\underline{80.91_{\pm0.10}}$

            & $37.68_{\pm1.79}$ & $50.65_{\pm1.94}$
            & $31.35_{\pm2.72}$ & $43.50_{\pm2.34}$ \\ 
{Dytox}    
            & $56.37_{\pm0.25}$ & $73.62_{\pm0.33}$
            & $47.60_{\pm0.39}$ & $67.58_{\pm0.26}$

            & $29.55_{\pm0.37}$ & $52.80_{\pm0.34}$
            & $23.62_{\pm0.50}$ & $47.83_{\pm0.57}$ \\ 
DRC    
            & $68.52_{\pm0.21}$ & $\underline{80.09_{\pm0.16}}$ 
            & $66.79_{\pm0.23}$ & $79.74_{\pm0.19}$

            & $35.18_{\pm0.63}$ & $54.79_{\pm0.82}$
            & $\underline{35.25_{\pm0.81}}$ & $55.26_{\pm0.85}$ \\ 
{HILAND}    
            & $56.18_{\pm1.07}$ & $68.76_{\pm0.89}$
            & $44.83_{\pm0.32}$ & $57.97_{\pm0.62}$

            & $\underline{40.33_{\pm1.68}}$ & $56.44_{\pm1.72}$
            & $34.22_{\pm1.79}$ & $50.22_{\pm1.80}$ \\ 
{BEEF}    
            & $61.96_{\pm0.15}$ & $73.76_{\pm0.63}$ & $57.16_{\pm0.83}$ & $70.26_{\pm0.20}$ & $33.11_{\pm0.89}$ & $52.49_{\pm0.49}$ & $27.79_{\pm0.85}$ & $48.07_{\pm0.82}$ \\
{DSGD}    
            & $62.76_{\pm0.37}$ & $74.48_{\pm0.49}$ & $58.49_{\pm0.58}$ & $70.96_{\pm0.55}$ & $35.23_{\pm0.33}$ & $53.74_{\pm0.35}$ & $30.12_{\pm0.57}$ & $50.29_{\pm0.81}$ \\
{CREATE}    
            & $64.86_{\pm0.52}$ & $75.95_{\pm0.41}$ & $60.40_{\pm0.29}$ & $72.36_{\pm0.48}$ & $36.66_{\pm0.44}$ & $54.55_{\pm0.36}$ & $33.84_{\pm0.27}$ & $52.54_{\pm0.11}$ \\
{VBM-TCIL}    
            & $68.14_{\pm0.08}$ & $78.02_{\pm0.03}$ & $67.16_{\pm0.09}$ & $76.83_{\pm0.00}$ & $37.12_{\pm0.23}$ & $55.17_{\pm0.29}$ & $35.12_{\pm0.58}$ & $53.73_{\pm0.68}$ \\

\rowcolor{lightpurple}
Ours 
            & $\underline{69.59_{\pm0.33}}$ & $\mathbf{80.34_{\pm0.18}}$ 
            & $\underline{69.81_{\pm0.41}}$ & $\mathbf{81.12_{\pm0.10}}$

            & $\mathbf{42.33_{\pm0.21}}$ & $\mathbf{62.27_{\pm0.08}}$
            & $\mathbf{37.21_{\pm0.89}}$ & $\mathbf{59.55_{\pm0.52}}$ \\
\bottomrule

\end{tabular}
}
\end{table}

\begin{table}[!h]
\centering
\caption{Performance comparison on Path16 and Skin8 dataset. ``Path16-X'' refers to the Path16 dataset organized in order X, ``Skin8-MX” denotes a memory size of X for the Skin8 dataset. The ``Average'' column aggregates the performance across four datasets.
``Last'' and ``Avg'' respectively represent Last-ACC and Avg-ACC. 
}

\label{eff-tab-path16-skin8}
\resizebox{1\linewidth}{!}{
\begin{tabular}{cccccccccccc}
\toprule
\multirow{2}{*}{Method}
& \multicolumn{2}{c}{Path16-I} & \multicolumn{2}{c}{Path16-II}
& \multicolumn{2}{c}{Skin8-M40} & \multicolumn{2}{c}{Skin8-M16} & \multicolumn{2}{c}{Average}\\
& Last & Avg & Last & Avg
& Last & Avg & Last & Avg & Last & Avg\\
\midrule
iCaRL & $65.52_{\pm0.34}$ & $78.49_{\pm0.36}$ 
      & $65.49_{\pm0.31}$ & $71.16_{\pm1.07}$
      & $41.17_{\pm1.42}$ & $62.15_{\pm0.40}$ 
      & $34.40_{\pm1.48}$ & $57.98_{\pm0.77}$
      & $46.36$ & $66.63$ \\
UCIR  & $56.75_{\pm1.23}$ & $75.16_{\pm0.28}$ 
      & $54.56_{\pm3.78}$ & $64.37_{\pm0.94}$
      & $\underline{43.87_{\pm2.19}}$ & $\underline{63.02_{\pm0.81}}$ 
      & $\underline{41.95_{\pm1.60}}$ & $\underline{61.94_{\pm1.51}}$
      & $47.55$ & $64.78$ \\
PODNet & $54.70_{\pm1.77}$ & $72.52_{\pm0.68}$ 
       & $56.16_{\pm0.27}$ & $65.33_{\pm0.30}$
       & $32.47_{\pm0.31}$ & $58.03_{\pm0.09}$ 
       & $30.95_{\pm2.56}$ & $56.27_{\pm0.16}$
       & $41.91$ & $62.40$ \\
DynaER & $\underline{69.45_{\pm0.75}}$ & $\underline{79.48_{\pm0.54}}$ 
       & $\underline{69.78_{\pm0.69}}$ & $\underline{74.54_{\pm0.30}}$
       & $39.54_{\pm3.36}$ & $60.18_{\pm1.46}$ 
       & $36.15_{\pm1.18}$ & $58.67_{\pm0.99}$
       & $47.93$ & $\underline{66.99}$ \\
WA     & $60.61_{\pm0.35}$ & $76.85_{\pm0.24}$ 
       & $57.99_{\pm1.57}$ & $66.59_{\pm0.82}$
       & $30.44_{\pm0.45}$ & $54.71_{\pm0.13}$ 
       & $24.82_{\pm0.60}$ & $50.59_{\pm0.33}$
       & $41.60$ & $61.55$ \\
DER++  & $58.87_{\pm2.81}$ & $75.15_{\pm0.96}$ 
       & $56.86_{\pm2.43}$ & $67.21_{\pm1.40}$
       & $29.73_{\pm2.23}$ & $57.89_{\pm1.67}$ 
       & $25.34_{\pm2.22}$ & $54.18_{\pm1.45}$
       & $44.06$ & $63.90$ \\
MORE   & $66.05_{\pm0.23}$ & $75.84_{\pm0.67}$ 
       & $62.50_{\pm0.08}$ & $69.01_{\pm1.23}$
       & $39.72_{\pm1.45}$ & $57.98_{\pm0.69}$ 
       & $35.16_{\pm2.06}$ & $55.95_{\pm1.31}$
       & $51.56$ & $64.19$ \\

{Dytox}       & $56.62_{\pm0.54}$ & $71.55_{\pm0.49}$ 
            & $56.80_{\pm0.62}$ & $66.08_{\pm1.37}$ 
            & $32.89_{\pm0.93}$ & $56.61_{\pm1.17}$ 
            & $25.24_{\pm1.29}$ & $49.69_{\pm1.21}$
            & $41.09$ & $60.72$ \\ 
DRC         & $61.39_{\pm0.71}$ & $76.71_{\pm0.33}$ 
            & $64.46_{\pm1.18}$ & $69.62_{\pm0.31}$
            & $39.26_{\pm1.68}$ & $59.92_{\pm0.63}$ 
            & $34.89_{\pm3.23}$ & $56.25_{\pm0.22}$
            & $50.72$ & $66.55$ \\
{HILAND}       & $60.44_{\pm1.17}$ & $75.61_{\pm0.11}$ 
            & $60.53_{\pm0.04}$ & $68.18_{\pm0.17}$ 
            & $37.31_{\pm0.91}$ & $58.51_{\pm0.46}$ 
            & $29.99_{\pm0.56}$ & $54.16_{\pm0.38}$
            & $45.48$ & $61.23$ \\ 
{BEEF}    
            & $61.21_{\pm0.58}$ & $74.70_{\pm0.56}$ & $59.93_{\pm0.47}$ & $68.94_{\pm0.93}$ & $33.12_{\pm0.78}$ & $56.96_{\pm0.52}$ & $27.87_{\pm0.45}$ & $53.81_{\pm0.74}$ & $45.27$ & $62.37$ \\
{DSGD}    
            & $63.21_{\pm0.66}$ & $75.84_{\pm0.23}$ & $62.92_{\pm0.72}$ & $68.67_{\pm0.74}$ & $38.72_{\pm0.22}$ & $60.56_{\pm0.31}$ & $33.94_{\pm0.86}$ & $57.68_{\pm0.51}$ & $48.17$ & $64.03$ \\
{CREATE}    
            & $65.35_{\pm0.49}$ & $77.39_{\pm0.02}$ & $63.77_{\pm0.37}$ & $69.62_{\pm0.22}$ & $37.25_{\pm0.64}$ & $59.36_{\pm0.34}$ & $34.77_{\pm0.69}$ & $57.87_{\pm0.44}$ & $49.61$ & $64.96$ \\
{VBM-TCIL}    
            & $66.25_{\pm0.72}$ & $77.56_{\pm0.30}$ & $65.78_{\pm0.05}$ & $69.45_{\pm0.96}$ & $41.27_{\pm0.68}$ & $62.25_{\pm0.22}$ & $36.68_{\pm0.75}$ & $59.49_{\pm0.37}$ & $\underline{52.19}$ & $66.56$ \\

\rowcolor{lightpurple} 
Ours        & $\mathbf{73.25_{\pm0.18}}$ & $\mathbf{81.00_{\pm0.43}}$ 
            & $\mathbf{72.25_{\pm0.35}}$ & $\mathbf{75.51_{\pm0.85}}$
            & $\mathbf{49.93_{\pm1.73}}$ & $\mathbf{66.42_{\pm1.08}}$ 
            & $\mathbf{44.81_{\pm2.08}}$ & $\mathbf{63.08_{\pm0.31}}$
            & $\mathbf{57.40}$ & $\mathbf{71.16}$ \\

\bottomrule
\end{tabular}
}
\end{table}

On the CIFAR100 dataset, when the number of tasks ($T$) was set to 10 and 20, the results are shown in Figure~\ref{eff-fig}(a) and Figure~\ref{eff-fig}(b) and in the CIFAR100-10T and CIFAR100-20T columns of Table~\ref{eff-tab-cifar100-cub200}. Our method achieved the best performance in Avg-ACC at 80.34\% and 81.12\%, and second best in Last-ACC at 69.59\% and 69.81\%, respectively, with only a slight difference of 0.30\% and 0.28\% from the top-performing MORE method. 

{On the CUB200 dataset, when the number of tasks ($\text{T}$) was set to 10 and 20, the corresponding results are reported in Figure~\ref{eff-fig}(c) and Figure~\ref{eff-fig}(d) and in the CUB200-10T and CUB200-20T columns of Table~\ref{eff-tab-cifar100-cub200}. Our method consistently achieved the best performance in Last-ACC, reaching 42.33\% and 37.21\% for the 10-task and 20-task settings, respectively. In terms of Avg-ACC, our method obtained 62.27\% and 59.55\% under the same task configurations. Notably, in the CUB200-10T setting, our method outperformed the second-best approach by a margin of 2.00\% in Last-ACC, clearly demonstrating its effectiveness in continual learning on fine-grained datasets.}

On the Path16 dataset, to comprehensively evaluate the model's performance under various task orders, we randomly sets two different task orders. Each order represents a pathway of incremental learning, with each task corresponding to a specific subset within Path16. Specifically, task order I is CO $\to$ LU $\to$ LN $\to$ ST $\to$ OC $\to$ CP $\to$ BR, and task order II is CP $\to$ BR $\to$ OC $\to$ LN $\to$ ST $\to$ LU $\to$ CO. The results are shown in the Figure~\ref{eff-fig}(e) and Figure~\ref{eff-fig}(f) and in the Path16-I and Path16-II columns of the Table~\ref{eff-tab-path16-skin8}. The figures demonstrate that under both task orders of the Path16 medical dataset, our proposed method consistently outperforms the baseline methods. Specifically, under different task orders, the final performance (Last-ACC) of our proposed method stabilized at around 73\%, approaching the performance upper bound of Joint at 74.55\%. 
In task order I, which progresses from simpler tasks (CO and LU)
, the performance of all methods was nearly 100\% initially. As the LN and ST tasks were introduced, the performance of baseline methods generally declined, while our proposed method maintained the lead. 
In the later stages of incremental learning, from the ST task to the BR task, our method showed performance improvements on these specific tasks, which led to an overall increase in the average performance across all tasks. 
The results shows that on the Path16 dataset, DynaER performed second best, with Last-ACC and Avg-ACC reaching 69.45\% and 79.48\%, respectively, yet our method still showed performance improvements of 3.80\% and 1.52\% over DynaER, respectively. Similar findings were observed under task order II.

On the Skin8 dataset, with memory sizes of 40 and 16 and the number of tasks ($T$) set to 4, the results are displayed in Figure~\ref{eff-fig}(g) and Figure~\ref{eff-fig}(h), as well as in the columns Skin8-M40 and Skin8-M16 of Table \ref{eff-tab-path16-skin8}. These results show that under both memory size settings, our method consistently maintained the best performance. With a memory size of 40, Last-ACC reached 49.93\% and Avg-ACC reached 66.42\%, showing improvements of 6.06\% and 3.40\% over the second best method, UCIR, which had Last-ACC at 43.87\% and Avg-ACC at 63.02\%, respectively. The performance under both memory settings confirms that our method is not sensitive to the size of the memory, maintaining best performance.

Further comparing our proposed method with the similar method MORE, it can be observed that our method outperforms MORE on most of the incremental learning settings across the majority of datasets, while being comparable performance on a few settings. On medical image datasets, the inferior performance of MORE is due to a certain degree of overfitting when the sample size is small. For example, on the initial task of the Skin8 dataset, the accuracy of MORE on the training set reaches 97.79\%, while its test accuracy is only 86.71\%. Similar results are observed in other tasks as well for our method and MORE. Additionally, MORE uses a random selection method to select old samples for replay, which selects less representative samples comparing to the herding selection method~\cite{icarl} that we use when the sample set is small. 
In summary, in terms of average performance across the four datasets, our proposed method not only shows significant improvement compared to the MORE method but also demonstrates substantial gains over all baseline methods.

\begin{figure}[h]
    \centering
    \includegraphics[width = 0.8\linewidth]{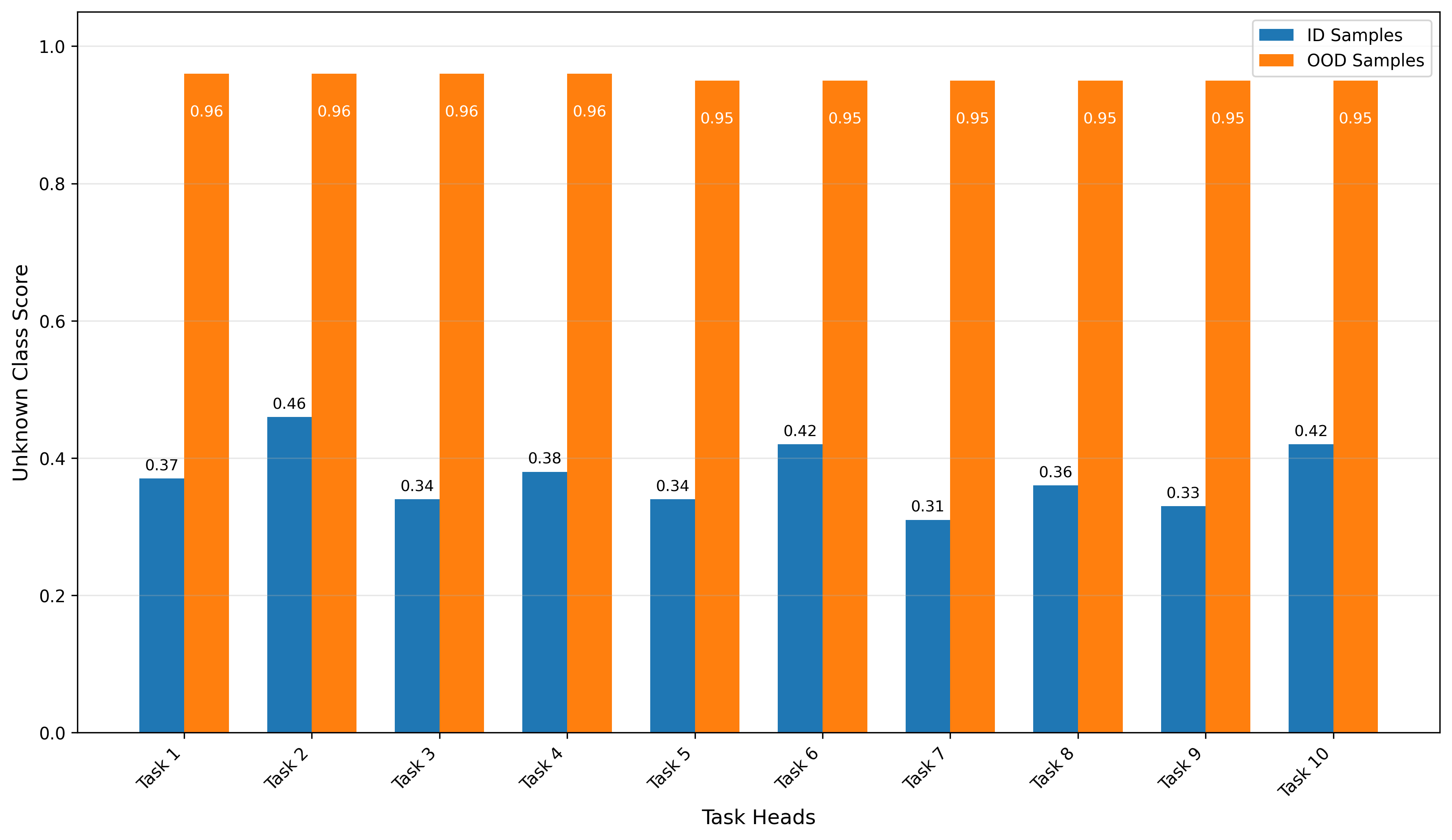}
    \caption{Comparison of unknown class scores between ID and OOD Samples under the 10-task setting on CIFAR100.}
    \label{id_ood_score_compare}
\end{figure}

Each task corresponds to a task-specific module (containing task-specific Batch Normalization and classification head). This configuration enables the probability output of the ``unknown class'' (aka `unknown class score') generated by each task-specific classification head to effectively distinguish between within-task and out-of-task samples. Here, we visualize average unknown class score from each task-specific sub-models for both within-task and out-of-task samples.
As shown in Figure~\ref{id_ood_score_compare}, each task-specific module consistently assigns lower unknown class scores to its corresponding in-task samples ($0.31 \sim 0.46$) compared to out-of-task samples ($0.95 \sim 0.96$). The visualization results demonstrate that our method effectively distinguishes between within-task and out-of-task samples through unknown class scores.



\subsection{Generalizabilization Assessment}

To investigate the generalization ability of the method proposed in this study, we utilized ResNet34 and ResNet50 architectures to further compare with selected baseline methods DynaER, iCaRL and DRC. {Meanwhile, we replaced the backbone with the lightweight architecture MobileNetV2~\cite{mobilenetv2} to evaluate the performance of our method in resource-constrained environments.}  It is noteworthy that although MORE performed well among the baselines, it is dependent on a Transformer network and is not adaptable to other ResNet architectures.  Therefore, MORE was excluded from this comparison, and only the best-performing and second-best performing method methods among the remaining baselines, DynaER, iCaRL and DRC, were selected for further analysis. The results are presented in Table~\ref{generalization}.

\begin{table}[!ht]
\centering
\caption{Comparison of the proposed method with baseline methods on three datasets with three backbone networks. Specifically, experiments on the Path16 dataset were conducted under task order I, on CIFAR100 under a division into 10 tasks, and on Skin8 with a memory setting of 40. ``Last'' and ``Avg'' represent Last-ACC and Avg-ACC, respectively.}
\label{generalization}
\resizebox{\linewidth}{!}{
\begin{tabular}{@{}llcccccccc@{}}
\toprule
\multirow{2}{*}{Dataset} & \multirow{2}{*}{Method }& \multicolumn{2}{c}{ResNet18} & \multicolumn{2}{c}{ResNet34} & \multicolumn{2}{c}{ResNet50} & \multicolumn{2}{c}{MobileNetV2} \\
\cmidrule(r){3-4} \cmidrule(r){5-6} \cmidrule(r){7-8} \cmidrule(r){9-10} 
 &  & Last & Avg & Last & Avg & Last & Avg & Last & Avg  \\
\midrule
\multirow{4}{*}{Path16} & iCaRL & $65.52_{\pm0.34}$ & $78.49_{\pm0.36}$  & $65.16_{\pm1.41}$ & $78.47_{\pm0.03}$ & $64.75_{\pm1.36}$ & $78.81_{\pm0.57}$ & $56.88_{\pm0.23}$ & $74.08_{\pm0.51}$ \\
 & DynaER & $69.45_{\pm0.75}$ & $79.48_{\pm0.54}$& $69.95_{\pm1.41}$ & $79.57_{\pm0.49}$ & $69.09_{\pm1.15}$ & $79.31_{\pm0.19}$  & $41.26_{\pm0.37}$ & $63.49_{\pm0.14}$ \\
 & DRC & $61.39_{\pm0.71}$ & $76.71_{\pm0.33}$ & $63.66_{\pm0.75}$ & $77.62_{\pm0.43}$ & $63.82_{\pm1.47}$ & $78.11_{\pm0.19}$ & $51.92_{\pm0.58}$ & $68.63_{\pm0.09}$ \\

 & \cellcolor{lightpurple} 
 Ours

 & \cellcolor{lightpurple} 
 $\bf{73.25}_{\pm0.18}$
 & \cellcolor{lightpurple} 
 $\bf{81.00}_{\pm0.43}$ 
 &\cellcolor{lightpurple} 
 $\bf{73.78}_{\pm1.01}$ 
 &\cellcolor{lightpurple} 
 $\bf{81.58}_{\pm0.57}$
 & \cellcolor{lightpurple}  
 $ \bf{75.11}_{\pm0.97}$ 
 & \cellcolor{lightpurple} 
 $\bf{82.30}_{\pm0.59}$ 
& \cellcolor{lightpurple}
$\bf{72.74}_{\pm0.42}$ 
& \cellcolor{lightpurple}
{$\bf{81.24}_{\pm0.26}$} \\
\midrule

\multirow{4}{*}{CIFAR100} & iCaRL &$57.75_{\pm0.35}$ &$75.74_{\pm0.18}$ & $60.08_{\pm0.24}$ & $76.89_{\pm0.17}$ & $60.03_{\pm 0.22}$ & $76.80_{\pm 0.17}$ & $55.00_{\pm0.18}$ & $73.47_{\pm0.33}$ \\
 & DynaER & $64.97_{\pm0.68}$ & $78.59_{\pm0.23}$ & $44.49_{\pm1.35}$ & $68.92_{\pm0.47}$ & $47.58_{\pm1.13}$ & $71.13_{\pm0.60}$ & $64.30_{\pm0.47}$ & $77.04_{\pm0.12}$ \\
 & {DRC} & $68.52_{\pm0.21}$ & $80.09_{\pm0.16}$ & $67.25_{\pm0.35}$ & $79.08_{\pm0.15}$ & $67.92_{\pm0.90}$ & $80.42_{\pm0.09}$  & $67.98_{\pm0.29}$ & $78.21_{\pm0.54}$ \\
 
 &\cellcolor{lightpurple} 
 Ours 
 &\cellcolor{lightpurple} 
 $\bf{69.59}_{\pm0.33}$
 &\cellcolor{lightpurple} 
 $\bf{80.34}_{\pm0.18}$
 &\cellcolor{lightpurple} 
 $\bf{72.10}_{\pm0.52}$ 
 &\cellcolor{lightpurple} 
 $\bf{79.66}_{\pm1.24}$ 
 &\cellcolor{lightpurple} 
 $\bf{73.52}_{\pm0.14}$ 
 &\cellcolor{lightpurple}
 $\bf{81.17}_{\pm1.10}$ 
& \cellcolor{lightpurple}
$\bf{71.54}_{\pm0.06}$ 
& \cellcolor{lightpurple}
$\bf{81.08}_{\pm0.39}$ \\

\midrule
 
 \multirow{4}{*}{Skin8} & iCaRL & $41.17_{\pm1.42}$ & $62.15_{\pm0.40}$ & $41.00_{\pm1.61}$ & $56.40_{\pm0.52}$ & $40.95_{\pm0.14}$ & $57.41_{\pm0.37}$ & $42.52_{\pm0.45}$ & $64.58_{\pm0.21}$ \\
 & DynaER & $39.54_{\pm3.36}$ & $60.18_{\pm1.46}$ & $41.19_{\pm 1.18}$ & $60.89_{\pm 0.50}$ & $36.56_{\pm 2.83}$ & $58.33 _{\pm 1.17}$ & $41.49_{\pm0.16}$ & $62.16_{\pm0.35}$ \\
 & {DRC} & $39.26_{\pm1.68}$ & $59.92_{\pm0.63}$ & $36.97_{\pm1.56}$ & $56.80_{\pm1.41}$ & $35.80_{\pm0.85}$ & $56.71_{\pm0.24}$ & $40.11_{\pm0.28}$ & $60.23_{\pm0.52}$ \\
 & \cellcolor{lightpurple} 
 Ours 
 & \cellcolor{lightpurple} 
 $\bf{49.93}_{\pm1.73}$ 
 & \cellcolor{lightpurple} 
 $\bf{66.42}_{\pm1.08}$ 
 & \cellcolor{lightpurple} 
 $\bf{51.07}_{\pm 2.09}$ 
 & \cellcolor{lightpurple} 
 $\bf{68.18}_{\pm 1.33}$ 
 & \cellcolor{lightpurple} 
 $\bf{52.87}_{\pm 0.92}$ 
 & \cellcolor{lightpurple} 
 $\bf{68.48}_{\pm 1.23}$ 
 & \cellcolor{lightpurple}
 $\mathbf{49.51}_{\pm0.11}$ 
 & \cellcolor{lightpurple}
 $\mathbf{66.19}_{\pm0.43}$ \\

\bottomrule
\end{tabular}
}
\end{table}

It can be observed that our method consistently maintained superior performance across different architectures. Moreover, as the network depth increased, our method demonstrated performance improvements on all three datasets. Specifically, when upgrading from ResNet18 to ResNet50, the Last-ACC for Path16, CIFAR100, and Skin8 datasets increased by 1.86\%, 3.93\%, and 2.94\% respectively, while the Avg-ACC increased by 1.30\%, 0.83\%, and 2.06\% respectively. By contrast, iCaRL's performance remained basically unchanged across different architectures. On the other hand, DynaER's performance showed inconsistent trends across different datasets, with fluctuating performances on Path16 and Skin8, and a significant decrease on CIFAR100, indicating a sensitivity to the network architecture employed. {For DRC, its performance also exhibits certain fluctuations across different architectures on the three datasets; however, under the same architecture, its performance remains inferior to ours, which further demonstrates the strong generalization capability of our method.}

{Furthermore, under lightweight backbone settings, our method consistently maintains superior performance across all three datasets. Specifically, when the backbone is replaced with MobileNetV2, the proposed method outperforms strong baseline iCaRL in terms of Last-ACC by 15.86\%, 16.54\%, and 6.99\% on the Path16, CIFAR100, and Skin8 datasets, respectively. Similarly, the Avg-ACC improves over iCaRL by 7.16\%, 7.61\%, and 1.61\%, respectively.
These results demonstrate that our approach remains robust and effective even under resource-constrained scenarios (e.g., reduced model parameters or lower computational complexity), highlighting its strong generalization capability. }

\subsection{Ablation Study}
The proposed method comprises three components, with ablation study results presented in Table~\ref{ablation_module}. The tabulated metrics represent average values from three independent random seeds (variances: 0.01-0.45). ``T.S.BN” denotes task-specific Batch Normalization (BN) - its removal implies shared pretrained BN layers across tasks, with only task-specific classification heads trainable during new task arrivals. ``unknown'' indicates adding an ``unknown'' class; its exclusion restricts task-ID prediction to maximum softmax scores without utilizing uncertainty scores.

\begin{table}[!h]
\centering
\caption{Ablation study of the proposed method. ``T.S.BN'' stands for task-specific Batch Normalization. ``unknown'' refers to the addition of an unknown class in the classification head, and ``Alignment'' denotes the OOD confidence alignment stage.}
\resizebox{\linewidth}{!}{
\begin{tabular}{ccc|cc|cccc}
\toprule
\multicolumn{3}{c|}{Components} & \multicolumn{2}{c|}{Path16} & \multicolumn{4}{c}{CIFAR100} \\
\cmidrule{1-3}
\cmidrule{4-5}
\cmidrule{6-9}
T.S.BN  & Unknown & Alignment  & Last-ACC & Avg-ACC & Last-TP & Avg-TP & Last-ACC & Avg-ACC \\
\midrule
\checkmark &            &                   &$18.61$ & $44.25$     
                           &  57.15 & 70.81& 56.89      &69.80

                                            \\ 
                                            
\checkmark & \checkmark &       &    $12.45$& $35.21$      

                          &$10.00$    &  $29.29$            &  $9.66$ & $28.29$ 
\\ 
                                            
& \checkmark & \checkmark   & $66.43$ & $77.39$    

                            &$53.84$&$69.86$               & $48.53$ & $62.70$   
\\                                          

\checkmark & \checkmark& \checkmark &  $\bf{73.25}$ & $\bf{81.00}$  
                     &$\bf{72.77}$&$\bf{83.40}$         & $\bf{69.59}$& $\bf{80.34}$ 
\\
\bottomrule
\end{tabular}
}
\label{ablation_module}
\end{table}

As shown in Table~\ref{ablation_module}, the first experiment (Row 1) demonstrates substantial performance degradation when replacing our OOD module with MSP~\cite{msp}, confirming the critical role of our OOD design.
In the second configuration (Row 2), removal of the alignment stage causes extreme behavior where Task 1's unknown class probability collapses to zero, inducing prediction bias. This validates the necessity of our alignment mechanism.
The third comparative study (Row 3) reveals that task-shared BN configuration exhibits significantly lower performance than task-specific BN. These results empirically verify that task-specific Batch Normalization (BN) enhances task-ID prediction performance through independent normalization of feature activations across individual tasks.

\begin{table}[h]
\centering
\caption{Performance with different OOD detection methods under the 10-task setting on CIFAR100.}
\label{different_metric_performance}
\resizebox{0.8\linewidth}{!}{ 
\begin{tabular}{ccccc}
\toprule
\textbf{Method} & \textbf{Last-TP} & \textbf{Avg-TP} & \textbf{Last-ACC} & \textbf{Avg-ACC}  \\
\midrule
MSP &  $57.15_{\pm0.35}$ & $70.81_{\pm0.63}$ & $56.89_{\pm0.38}$ & $69.80_{\pm0.60}$\\
MaxLogit &  $60.83_{\pm0.66}$  &$74.12_{\pm0.62}$&$60.11_{\pm0.58}$&$72.57_{\pm0.63}$ \\
ODIN & $59.24_{\pm0.76}$ &$72.28_{\pm1.00}$&$57.92_{\pm0.88}$&$67.71_{\pm1.48}$\\
Entropy  & $57.49_{\pm0.49}$&$71.09_{\pm0.63}$ &$57.24_{\pm0.53}$&$70.08_{\pm0.60}$ \\
Energy  &$60.82_{\pm0.64}$ &$74.10_{\pm0.60}$&$60.07_{\pm0.57}$&$72.51_{\pm0.62}$ \\
ReAct&$57.90_{\pm0.77}$ & $71.38_{\pm0.67}$&$57.61_{\pm0.79}$&$70.36_{\pm0.67}$ \\

\rowcolor{lightpurple} 
Unknown(Ours)  & $\bf{72.77}_{\pm0.14}$ & $\bf{83.40}_{\pm0.10}$&$\bf{69.59}_{\pm0.33}$ &$\bf{80.34}_{\pm0.18}$ \\
\bottomrule
\end{tabular}}
\label{ablation_ood}
\end{table}

To benchmark OOD capability, we conduct controlled ablation by retaining only T.S.BN (ablation version in Table~\ref{ablation_module} Row 1). Crucially, we extend evaluation beyond conventional softmax to multi-metric analysis including MSP~\cite{msp}, MaxLogit~\cite{maxlogit}, ODIN~\cite{odin}, Entropy, Energy~\cite{energy}, and ReAct~\cite{react}. As shown in Table~\ref{different_metric_performance}, our method outperforms the second-best approach by +11.94\% in Last-TP and +9.28\% in Avg-TP, demonstrating that our framework enhances OOD detection capability while improving TP performance.



\subsection{Analysis of Selecting Top-$k$ Classification Heads with the Lowest ``Unknown'' Scores}
\begin{figure}[h]
    \centering
    \includegraphics[width = 1\linewidth]{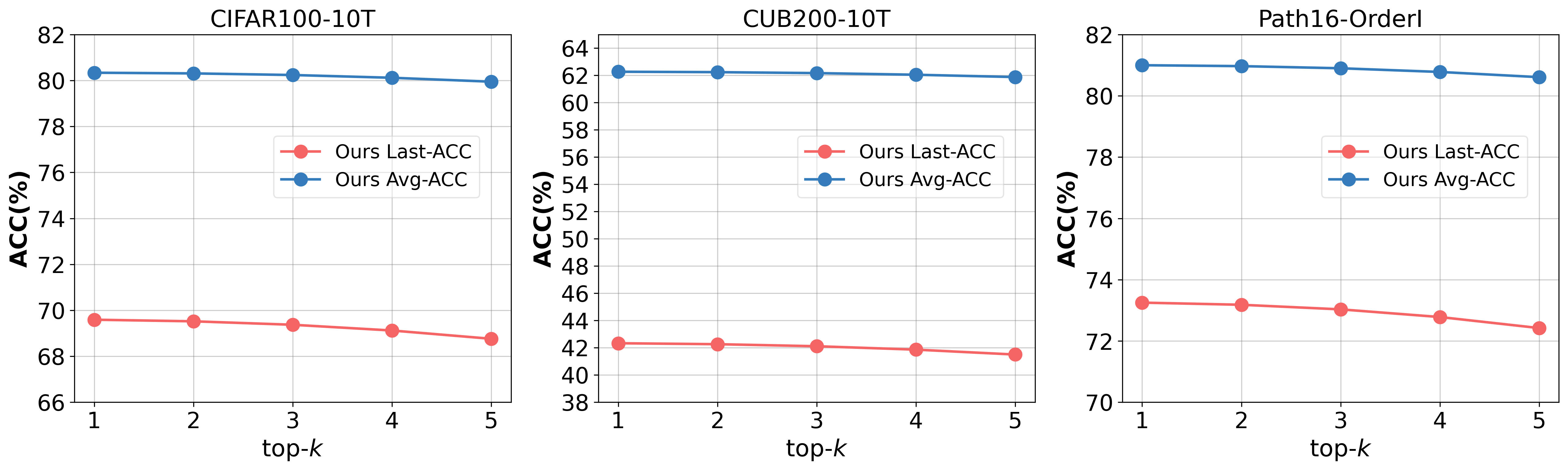}
    \caption{Impact of selecting top-k classification heads with the lowest ``unknown'' scores.}
    \label{sensitivity}
\end{figure}

We further evaluated a soft assignment mechanism that performs category prediction by aggregating the outputs of the top-k classification heads with the lowest ``unknown'' class scores on the CIFAR100-10T, CUB200-10T, and Path16-OrderI. As shown in Figure~\ref{sensitivity}, both Last-ACC and Avg-ACC exhibit slight declines as the number of selected classification heads increases from 1 to 5 on three datasets. This suggests that incorporating outputs from additional task heads tends to introduce noise from irrelevant tasks. In contrast, selecting only the most appropriate classification head enables the model to better exploit task-specific knowledge after task identification. These results demonstrate the effectiveness of our design choice to rely on a single best-matched task-specific classification head.

\color{black}

\subsection{Robustness of Task-ID Prediction}

The performance of our method largely depends on the accuracy of task-ID prediction. As the number of tasks increases, whether our method can maintain stable TP performance becomes a critical concern. Therefore, we conducted detailed experiments on the CIFAR100 dataset with different numbers of tasks (10, 20, 25, 50). 
The results are shown in Table~\ref{task_num_ours}. Last-TP denotes the final task-ID prediction accuracy after completing all phases, Avg-TP represents the average accuracy across phases; WP indicates the within-task classification accuracy given predicted task-ID (Last-WP for final accuracy, Avg-WP for phase-wise average); Last-ACC and Avg-ACC are standard incremental learning metrics. 



\begin{table}[h]
\centering
\caption{Performance with varying number of tasks on CIFAR100.}
\label{task_num_ours}
\resizebox{\linewidth}{!}{
\begin{tabular}{ccccccc}
\toprule
Task Count& Last-TP & Avg-TP & Last-WP & Avg-WP  & Last-ACC & Avg-ACC\\
\midrule
$T = 10$ &$72.77_{\pm0.14}$ & $83.40_{\pm0.10}$ & $95.63_{\pm0.40}$  &$96.34_{\pm0.17}$  &$69.59_{\pm0.33}$ &$80.34_{\pm0.18}$  \\
$T = 20$ &$70.83_{\pm0.41}$  & $82.34_{\pm0.13}$ & $98.26_{\pm0.10}$  & $98.14_{\pm0.06}$ &$69.81_{\pm0.41}$&$81.12_{\pm0.10}$ \\
$T = 25$  & $70.41_{\pm0.13}$ & $81.79_{\pm0.17}$ & $98.49_{\pm0.10}$  & $98.53_{\pm0.10}$& $69.35_{\pm0.06}$ & $80.58_{\pm0.24}$  \\
$T = 50$  & $68.52_{\pm0.16}$ & $81.04_{\pm0.20}$ &$99.46_{\pm0.02}$ &$99.29_{\pm0.21}$&$68.15_{\pm0.16}$&$80.46_{\pm0.18}$\\

\bottomrule
\end{tabular}
\label{ablation}
}
\end{table}

Table~\ref{task_num_ours} demonstrates that our method maintains stable incremental learning performance as task numbers increase, with Last-ACC stabilizing around 68\% and Avg-ACC remaining at approximately 80\%. This stability originates from sustained consistency in both TP and WP performance. Specifically, task-ID prediction performance shows no significant degradation across task scales, with Last-TP consistently maintained above 68\% and Avg-TP sustained above 81\%., which validates the effectiveness of our OOD design. Meanwhile, within-task prediction accuracy not only consistently exceeds 95\% but improves with task accumulation: Last-WP rises from 95.63\% to 99.46\% and Avg-WP from 96.34\% to 99.29\%. This divergence confirms that task-ID prediction constitutes the primary challenge, whereas within-task prediction becomes substantially more reliable when task identification is correct.

\begin{table}[h]
\centering
\caption{Performance comparison between the proposed method and baselines under varying number of tasks on CIFAR100.}
\label{task_num_baseline}
\resizebox{1.0\linewidth}{!}{
\begin{tabular}{ccccccccc}
\toprule
\multirow{2}{*}{Method} & \multicolumn{2}{c}{$T$ = 10} & \multicolumn{2}{c}{$T$ = 20} & \multicolumn{2}{c}{$T$ = 25} & \multicolumn{2}{c}{$T$ = 50} \\
& Last-ACC & Avg-ACC & Last-ACC & Avg-ACC & Last-ACC & Avg-ACC & Last-ACC & Avg-ACC\\
\midrule
UCIR        &$65.99_{\pm0.60}$   &$76.63_{\pm0.41}$  &$55.65_{\pm0.60}$ &$73.95_{\pm0.19}$ &$54.89_{\pm0.43}$&$74.18_{\pm0.44}$&$43.85_{\pm2.56}$ & $69.69_{\pm1.17}$  \\
DynaER        &$64.97_{\pm0.68}$   &$78.59_{\pm0.23}$ &$42.65_{\pm0.58}$ &$67.74_{\pm0.31}$ &$45.68_{\pm0.98}$&$70.28_{\pm0.09}$ &$47.54_{\pm0.13}$&$68.84_{\pm0.03}$\\
DER++       & $66.69_{\pm0.54} $ &$79.90_{\pm0.34}$&$60.21_{\pm0.99}$ &$77.10_{\pm0.10}$& $60.17_{\pm0.57}$ & $77.58_{\pm0.23}$ &$52.83_{\pm1.15}$ & $71.92_{\pm0.18} $ \\
\rowcolor{lightpurple} 
Ours       & $\bf{69.59}_{\pm0.33}$  &$\bf{80.34}_{\pm0.18}$&$\bf{69.81}_{\pm0.41}$  & $\bf{81.12}_{\pm0.10}$& $\bf{69.35}_{\pm0.06}$&$\bf{80.58}_{\pm0.24}$&$\bf{68.15}_{\pm0.16}$&$\bf{80.46}_{\pm0.18}$ \\
\bottomrule
\end{tabular}
}
\end{table}

We conducted comparative evaluations with three top-performing baselines—UCIR, DynaER, and DER++—under identical experimental conditions. As shown in Table~\ref{task_num_baseline}, in small-scale scenarios with T=10 tasks, our method demonstrates a marginal advantage where DER++’s Last-ACC trails by merely 2.90\%. However, as task quantities escalate, baseline methods exhibit progressive degradation: DER++’s Last-ACC plummets from 66.69\% to 52.83\% and Avg-ACC declines from 79.90\% to 71.92\%. In stark contrast, our approach maintains stable performance across all scales, thereby confirming its task-number robustness for real-world deployment scenarios.

\subsection{Analysis of Computational Complexity}  
we analyzed the training and inference time per task under the CIFAR100 10-task setting for both our method strong baseline DynaER~\cite{DER}. As shown in Table~\ref{tab:training_time}, the primary factor contributing to the increased training time of our method is the OOD alignment stage. However, this stage does not exhibit a pronounced linear growth as the number of tasks increases, since it only requires fine-tuning the final classifier heads. Compared with the strong baseline DynaER, the overall computational overhead of our method remains within an acceptable range. Regarding inference efficiency (shown in Table~\ref{tab:inference_time}), the testing time of our method does not increase rapidly with the number of tasks but instead remains relatively stable, making it suitable for real-world deployment scenarios where fast inference is required. 

\begin{table}[t]
    \centering
    \caption{Training time of baseline method and our method on CIFAR100 under 10-task setting.}
    \resizebox{\linewidth}{!}{
    \begin{tabular}{lccccccccccc}
        \toprule
        \multirow{2}{*}{\textbf{Method}}  & \multicolumn{10}{c}{\textbf{Time (minute)}}& \multirow{2}{*}{\textbf{Epoch}} \\
        \cmidrule(lr){2-11}
        &  \textbf{Task 1} & \textbf{Task 2} & \textbf{Task 3} & \textbf{Task 4} & \textbf{Task 5} & \textbf{Task 6} & \textbf{Task 7} & \textbf{Task 8} & \textbf{Task 9} & \textbf{Task 10}& \\
        \midrule
        DynaER (Stage 1)  & 15.53 & 76.87 & 59.67 & 90.35 & 75.59 & 81.91 & 75.72 & 88.65 & 103.14 & 118.47 & 200 \\
        DynaER (Stage 2)  & 0.00 & 30.85 & 21.60 & 39.14 & 19.02 & 16.47 & 19.08 & 22.01 & 25.46 & 29.12 & 50 \\
        DynaER (Total)  & 15.53 & 102.72 & 76.26 & 124.50 & 89.61 & 93.38 & 89.80 & 105.67 & 123.59 & 142.59 & - \\
        \midrule
        Ours (Stage 1)  & 12.66 & 21.00 & 24.46 & 27.48 & 24.41 & 30.88 & 32.40 & 27.81 & 27.37 & 27.91 & 200 \\
        Ours (Stage 2)  & 0.00 & 17.05 & 8.35 & 22.07 & 16.34 & 18.63 & 20.53 & 19.65 & 22.81 & 27.37 & 50 \\
        Ours (Total)  & 12.66 & 38.05 & 32.81 & 49.55 & 40.75 & 49.51 & 52.94 & 47.46 & 50.19 & 55.28 & - \\

        \bottomrule
    \end{tabular}}
    \label{tab:training_time}
\end{table}

\begin{table}[!t]
    \centering
    \caption{Inference time of baseline method and our method for each test image in CIFAR100 under 10-task setting.}
    \label{tab:inference_time}
    \resizebox{\linewidth}{!}{
    \begin{tabular}{lcccccccccc}
        \toprule
        \multirow{2}{*}{\textbf{Method}}  & \multicolumn{10}{c}{\textbf{Inference time (millisecond)}}  \\ 
        
        \cmidrule(l){2-11} 
        & \textbf{Task 1} & \textbf{Task 2} & \textbf{Task 3}& \textbf{Task 4} & \textbf{Task 5} & \textbf{Task 6} & \textbf{Task 7} & \textbf{Task 8} & \textbf{Task 9} & \textbf{Task 10} \\
         
        \midrule
        
        DynaER  & 10.25 & 10.44 & 11.90 & 12.28 & 13.36 & 14.27 & 15.65 & 17.00 & 17.85 & 19.36 \\
        Ours  & 5.11 & 5.12 & 5.12 & 5.13 & 5.15 & 5.16 & 5.18 & 5.20 & 5.20 & 5.21 \\

        \bottomrule
    \end{tabular}
    }
\end{table}

\begin{figure}[h]
    \centering
    \includegraphics[width = 0.6\linewidth]{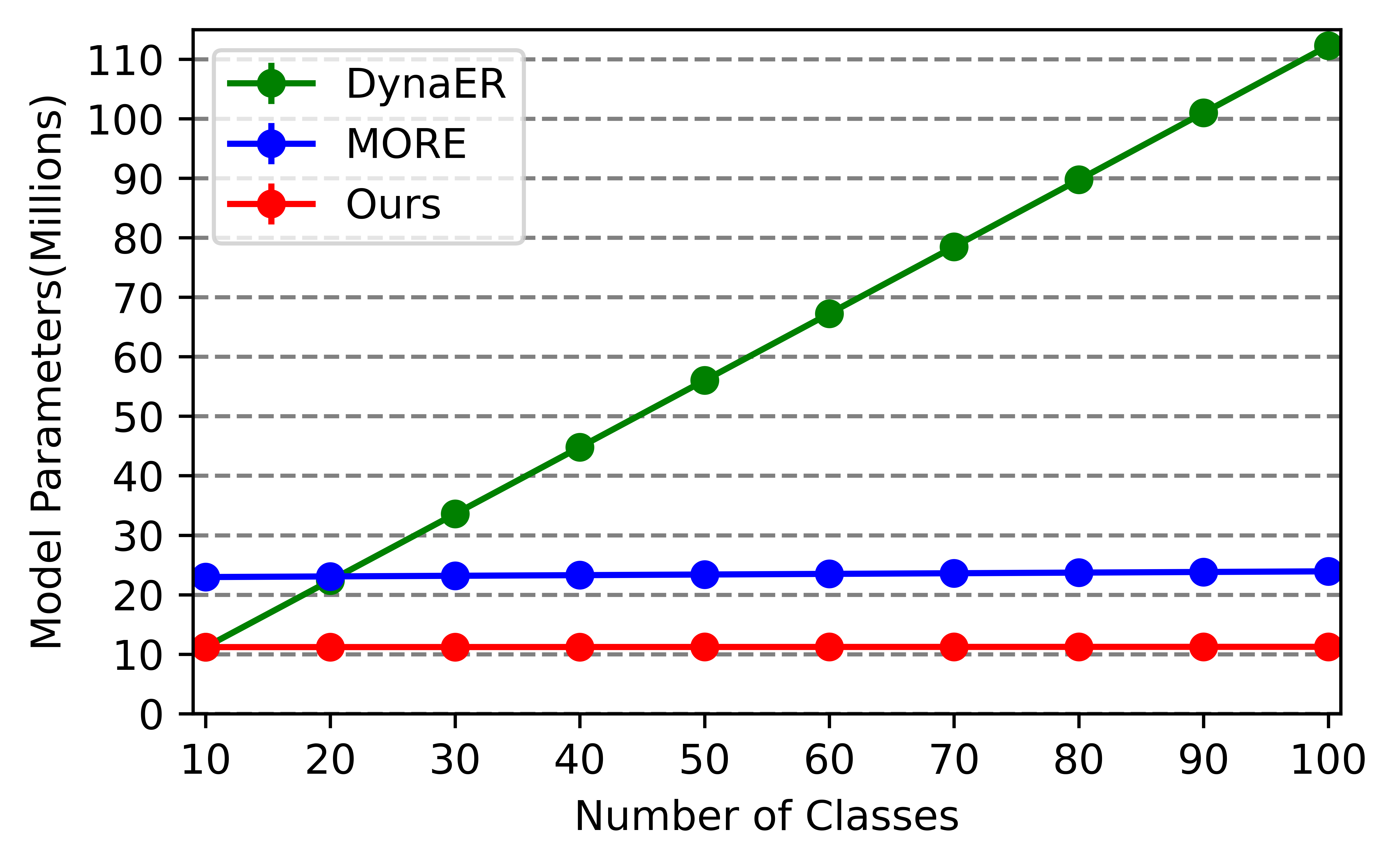}
    \caption{Comparison of parameter increase between the proposed method and the best-performing baseline methods, DynaER and MORE, on the CIFAR100 dataset with 10 tasks.}
    \label{model_params}
\end{figure}

We further analyze the parameter growth trends of our method compared with DynaER and MORE~\cite{more} under the CIFAR100 10-task setting. As shown in Figure~\ref{model_params}, our method achieves significantly better parameter efficiency in the CIL setting.
Specifically, DynaER introduces 11.2M additional parameters per task, while MORE adds approximately 70K parameters for each new task, both resulting in considerable model expansion and increased computational overhead. In contrast, our method requires only 15K additional parameters per task while achieving the best average performance across four datasets, highlighting its ability to balance model compactness and continual learning performance effectively.

To further evaluate scalability to long task sequences, we measure training time, inference time, and model parameters under varying numbers of tasks on CIFAR100. As reported in Table~\ref{tab:all}, when the number of tasks increases from 10 to 50, our method consistently requires less training time, lower inference time, and fewer parameters than DynaER. These results demonstrate that our approach maintains strong scalability even in long task sequences.

\begin{table}[hb]
    \centering
    \caption{Comparison of training time (minute), inference time (millisecond), and model parameters (million) of final task across different numbers of tasks under CIAFR100.}
    \label{tab:all}
    \resizebox{1\linewidth}{!}{
    \begin{tabular}{lccccccccc}
        \toprule
        \multirow{2}{*}{ Method } &\multicolumn{3}{c}{{T=10}} &\multicolumn{3}{c}{{T=20}}  &\multicolumn{3}{c}{{T=50}} \\

        \cmidrule(lr){2-4} \cmidrule(lr){5-7} \cmidrule(lr){8-10} 
        
        & Training time  & Inference time  & Model size  & Training time  & Inference time  & Model size & Training time  & Inference time  & Model size \\ 
        
        \midrule

        DynaER (Stage 1)  & 118.47 & \multirow{3}{*}{19.36} & \multirow{3}{*}{112.28} & 187.18 & \multirow{3}{*}{25.33} & \multirow{3}{*}{223.80} & 549.31 & \multirow{3}{*}{66.84} & \multirow{3}{*}{559.50} \\
        DynaER (Stage 2)  & 29.12 &  &  &  56.83 &  &  & 133.64 &  &  \\
        DynaER (Total)  & 142.59 &  &  &  286.47 &  &  & 682.95 &  &  \\
        \midrule
        Ours (Stage 1)  & 27.91 & \multirow{3}{*}{5.21} & \multirow{3}{*}{11.24} & 18.87 & \multirow{3}{*}{5.29} & \multirow{3}{*}{11.29} &  15.85 & \multirow{3}{*}{6.33} & \multirow{3}{*}{11.45} \\
        Ours (Stage 2)  & 27.37 &  &  & 41.78 &  &  &  95.63 &  & \\
        Ours (Total)  & 55.28 &  &  & 60.65 &  &  &  111.48 &  &  \\

        \bottomrule
    \end{tabular}}
\end{table}
\color{black}

\section{Conclusion and Limitations}
To address the catastrophic forgetting problem and achieve a better balance between stability, plasticity, and parameter growth, we proposed a class incremental learning method based on task-specific batch normalization and classification heads combined with OOD detection. This method has reached the state-of-the-art performance on the medical image datasets Skin8, Path16, and the natural image dataset CIFAR100 and CUB200. The generalization experiments further confirmed the adaptability of our method across models with different scales, and ablation studies validated the effectiveness of each component.

However, our method still has several limitations. First, it requires a memory buffer for replaying old-task samples, which introduces additional memory overhead. To address this issue, feature replay~\cite{slca} instead of sample replay could be adopted  by using generated features from previous tasks as OOD samples to train the OOD detection capability of task-specific classification heads, which is more storage-efficient.

Second, since Batch Normalization is a standard component in convolutional neural networks, our strategy of progressively expanding BN layers for incoming tasks is broadly applicable to BN-based architectures. For Transformer-based models such as Vision Transformer~\cite{vit}, task-specific Layer Normalization could be introduced to replace the original normalization layers. Additionally, other normalization strategies, such as Instance Normalization or Group Normalization, could be explored as task-specific alternatives to BN.

Finally, although our method improves task-ID prediction through the OOD Detection Alignment Stage by aligning the confidence outputs of task-specific classification heads via balanced memory replay, it introduces additional computational overhead because all existing classifier heads must be retrained during each alignment step. {This computational overhead could be alleviated by leveraging stored features of previous samples~\cite{darker} or generated pseudo-features~\cite{slca} to fine-tune all task-specific adapters, instead of retaining raw images. In this way, each sample would no longer need to pass through the full adapted backbone, thereby avoiding the associated computational cost. Instead, only the lightweight task-specific classification head would be passed through, while still maintaining effective OOD alignment.}
\color{black}

\section{Acknowledgements}
Funding: This work was supported by National Natural Science Foundation of China (grant No. 62071502).

\bibliographystyle{elsarticle-num}
\bibliography{refs}
\end{document}